# Intrinsic static/dynamic triboelectric pressure sensor for continuous and event-triggered control

Kequan Xia[1], Song Yang[1], Jianguo Lu[2], Min Yu[1*]

[1]Department of Mechanical Engineering, Imperial College London, SW7 2AZ, London, United Kingdom;

[2]State Key Laboratory of Silicon and Advanced Semiconductor Materials, School of Materials Science and Engineering, Zhejiang University, Hangzhou 310058, China.

*Corresponding author: m.yu14@imperial.ac.uk.

*Abstract*—Conventional pressure sensors often integrate two distinct mechanisms to detect static and dynamic stimuli, hindering the development of high-fidelity human–machine interfaces. Here, we present an intrinsic static–dynamic triboelectric sensor (iSD Sensor) capable of reliably perceiving both continuous static pressure and transient mechanical shocks through a DC/AC signal decoupling strategy. By pairing hydrophobic expanded polytetrafluoroethylene (ePTFE) with elastic conductive sponge, a pressure-adaptive triboelectric interface is formed, where microscale and large-scale separations enable static and dynamic pressure sensing, respectively. Furthermore, by employing a charge excitation strategy, the device delivers enhanced voltage outputs—over 25× in static and 15× in dynamic modes. Combined with a 3D gradient conductive sponge structure, the sensor achieves multi-region sensitivities of 34.7 $V \cdot kPa^{-1}$ (static) and 48.4 $V \cdot kPa^{-1}$ (dynamic) under low pressure (<1.8 kPa), and a detection limit as low as 6.13 Pa. By perceiving continuous static pressure and transient shocks applied by the human hand, the iSD Sensor enables robotic arm control via proportional grasping and dynamic, trigger-based sign language communication. This work advances high-sensitivity, self-powered pressure sensors toward intelligent, closed-loop human–machine interaction.

## 1. Introduction

The ability to perceive tactile cues—particularly pressure stimuli—is essential for enabling robotic systems to interact intelligently with humans and complex, unstructured environments [1]. Tactile feedback plays a pivotal role in regulating grip strength, modulating motion trajectories, and enabling real-time environment-aware behaviors in a broad range of applications, including prosthetic limbs [4, 5], collaborative robots [2, 3], and human-machine interaction (HMI) [6, 7]. Inspired by the spatial and temporal acuity of human mechanoreceptors, researchers have developed flexible pressure sensors as foundational components of electronic skin (e-skin), aiming to replicate the human ability to sense contact force [8], texture [9], and stiffness identification [10]. Despite notable progress in enhancing sensitivity, flexibility, and stretchability, most existing pressure sensors suffer from two critical limitations: dependence on external power supplies and limited sensing modality. Capacitive and piezoresistive sensors, though widely used due to their high sensitivity and tunability, require continuous electrical excitation and signal amplification, which increases power consumption and constrains system-level integration, especially in mobile or battery-constrained platforms [11, 12]. Furthermore, these sensors are prone to performance degradation such as signal drift, hysteresis, and mechanical fatigue under prolonged use or large deformation, thereby limiting their long-term stability and robustness. A more fundamental challenge lies in their inability to simultaneously resolve static and dynamic pressure signals. In many sensor designs, the mechanisms responsible for generating measurable output favor either transient stimuli (e.g., tapping, vibrations) or sustained loads (e.g., static holding), but not both. For instance, piezoelectric sensors inherently produce transient signals and rapidly decay under steady-state pressure [13, 14], while capacitive sensors [15], though capable of detecting sustained forces, exhibit delayed or attenuated response to fast dynamic input. The intrinsic decoupling of static and dynamic pressure modes remains a fundamental barrier to achieving biologically inspired, continuous tactile feedback for robotic manipulation. Addressing this challenge requires the development of compact, self-powered pressure sensors that operate without external power while simultaneously distinguishing static and dynamic stimuli in real time—enabling closed-loop control, intelligent grasp modulation, and tactile communication such as sign-language encoding.

Triboelectric nanogenerators (TENGs) have emerged as a promising solution for self-powered sensing and energy harvesting, converting mechanical stimuli into electrical signals through the combined effects of triboelectrification and electrostatic induction, without requiring external power [16-25]. Since their introduction in 2012, TENGs have demonstrated strong potential in harvesting energy from various mechanical sources—including human motion [26-28], wind [29, 30], vibrations [31-34], and water waves [35-37]—enabling broad applications in robotic systems and autonomous sensing platforms. Triboelectric sensors for robotics can generally be classified into two major research directions: (1) fundamental studies exploring strain perception mechanisms arising from contact or deformation, and (2) application-driven developments targeting practical functions such as intelligent grasping, object and texture recognition, multi-modal tactile sensing, and human–machine interaction [38]. Triboelectric pressure sensors have shown considerable promise in HMI systems by enabling real-time, self-powered tactile communication. For example, Chen et al. developed a flexible, self-powered HMI patch based on a sliding-mode PTFE–copper triboelectric pair, which enables reliable finger-motion sensing via joint deformation for the remote control of a soft robotic manipulator [39]. Zhu et al. developed a smart glove integrating triboelectric finger and palm sensors with sliding and bending modes, enabling multi-directional motion detection and haptic feedback for immersive, self-powered human–machine interaction in VR/AR scenarios [40]. Immediately, Zhu et al. developed a modular soft glove integrating TENG-based tactile and strain sensors with pneumatic and thermal haptic feedback in multifunctional 'Tactile+' units, enabling real-time bidirectional communication between humans and machines through compact, material-driven sensor–actuator fusion [41]. Recently, Chen et al. proposed a soft finger case incorporating 3D multi-modal tactile sensing along with thermal and haptic feedback, facilitating more natural and immersive interaction between robots and virtual environments [42]. Although triboelectric HMI systems have seen significant progress, most remain constrained by unstable static pressure sensing, dependence on bulky external modules, and the absence of standardized integration strategies—factors that limit their applicability in compact, real-time robotic platforms. While triboelectric pressure sensors have demonstrated excellent dynamic sensing performance, their capability for static pressure detection is still hindered by rapid charge dissipation and the lack of sustained signal output. In addition, current studies often rely on non-standardized, device-specific measurement setups, leading to inconsistent evaluation of static and dynamic responses across different platforms [43-45]. Although hybrid approaches that incorporate capacitive or piezoresistive components have been explored to address this issue, they often increase system complexity, fabrication cost, and the risk of signal interference [46, 47]. To overcome these limitations, our recent work proposed an all-foam triboelectric sensor featuring a standardized DC/AC measurement strategy, capable of directly generating high-resolution static and dynamic pressure signals without relying on hybrid mechanisms or external signal-processing modules [48]. Leveraging both static and dynamic pressure signals for robotic control offers a meaningful step toward more intuitive, adaptive, and responsive human–machine interaction.

In this work, we present an intrinsic static–dynamic triboelectric sensor (iSD Sensor) capable of high-resolution detection of both sustained and transient pressure stimuli, achieved through a DC/AC signal decoupling strategy implemented via a combined electrometer–oscilloscope. In detail, the iSD Sensor integrates expanded polytetrafluoroethylene (ePTFE) and elastic conductive sponge as triboelectric layers, with aluminum foil and conductive sponge serving as the electrodes. The integration of hydrophobic ePTFE with elastic conductive sponge yields a pressure-adaptive triboelectric interface, in which distinct contact–separation behaviors—microscale for static and large-scale for dynamic—enable intrinsic dual-mode pressure sensing. Furthermore, a charge excitation (CE) strategy is employed to actively boost interfacial charge density, leading to more than 25-fold and 15-fold enhancements in static and dynamic voltage outputs, respectively. The iSD

Sensor exhibits distinct multi-region sensitivity profiles in both sensing modes. For static pressure, sensitivities of 2.6 V·kPa$^{-1}$, 0.7 V·kPa$^{-1}$, and 0.2 V·kPa$^{-1}$ are observed in the low (0.6–9.8 kPa), medium (12.2–24.5 kPa), and saturation (>27.5 kPa) ranges, respectively. For dynamic pressure, corresponding sensitivities of 9.8, 2.3, and 0.1 V·kPa$^{-1}$ are achieved across 0–5.5 kPa, 7.3–12.2 kPa, and beyond 15.3 kPa. The 3D stacked gradient structure significantly improves contact dynamics and sensing resolution in the low-pressure regime, yielding peak sensitivities of 34.7 V·kPa$^{-1}$ and 48.4 V·kPa$^{-1}$ for static and dynamic modes, respectively, within pressures below 1.8 kPa, along with a minimum detection limit of 6.13 Pa. The iSD Sensor detects both static and dynamic pressures applied by the human hand and, through a DC/AC signal decoupling strategy, enables robotic arm control—using static inputs for proportional object grasping and dynamic signals for trigger-based sign language transmission. This work presents the first fully integrated, self-powered tactile interface capable of intrinsically distinguishing and leveraging static and dynamic pressure signals for direct robotic actuation, marking a paradigm shift toward dual-mode, application-ready platforms for next-generation robotic control and assistive human–machine interaction.

# 2. Experiment

### 2.1. Fabrication process of iSD Sensor device.

The iSD Sensor is fabricated through a modular and layer-by-layer assembly process, combining elastic supporting structural components and triboelectric layers to enable intrinsic static/dynamic pressure sensing. The adhesive double-sided conductive sponge (thickness: 1 mm) and the ePTFE film (thickness: 0.5 mm) serve as a triboelectric pair, while the conductive sponge and the aluminum foil (size: 4 cm × 4 cm) function as a coupled electrode structure. The acrylic sheet serves as the substrate supporting both the triboelectric layers and the electrodes, while an elastic sponge interlayer is introduced to modulate the maximum separation gap between the triboelectric layers, thereby enabling effective dynamic pressure sensing. The ePTFE film, with its high electronegativity, porous microstructure, and mechanical flexibility, serves as an effective triboelectric layer that enhances charge separation efficiency and ensures conformal contact under both static and dynamic pressure stimuli. The adhesive conductive sponge in this design serves both as a flexible support and a triboelectric functional layer, enabling stable assembly, efficient pressure sensing with reliable conductivity, while also providing a manufacturable foundation for the 3D gradient structural design. Besides, the 3D gradient design consists of five conductive sponge layers with dimensions of 4.5 cm × 4.5 cm, 3.6 cm × 4.5 cm, 2.7 cm × 4.5 cm, 1.8 cm × 4.5 cm, and 0.9 cm × 4.5 cm, each with a uniform thickness of 2 mm. The iSD Sensor designed for robotic control features a compact 5 cm × 5 cm form factor, allowing easy integration into various end-effectors or wearable platforms.

### 2.2. Characterization and measurement.

To evaluate the output characteristics of the iSD Sensor, a precisely configured mechanical excitation system—comprising a signal generator, power amplifier, exciter, and a reference pressure sensor—was employed. This system allows for flexible adjustment of motion parameters and provides accurate control over excitation waveforms such as sinusoidal and square waves. The output voltage signals of the iSD Sensor were measured using a Keithley 6517B electrometer. For dynamic pressure sensing, the sensing signals were characterized in real time via the AC mode of a TBS 2000 oscilloscope, while static pressure signals were displayed through the DC mode of the TBS 2000 oscilloscope. The applied pressure on the iSD Sensor was quantified using a digital force measurement system, while the corresponding mass was measured with a high-precision electronic balance. The self-developed signal modulation and acquisition module facilitates portable control signal acquisition, enabling accurate grip manipulation and effective sign language communication in the robotic arm.

# 3. Results and discussion

### 3.1. Self-powered control architecture based iSD Sensor for multifunctional robotic manipulation.

Fig. 1(a) shows a conceptual framework of a self-powered control system for robotic manipulation, enabled by intrinsic static and dynamic pressure sensing. A key innovation of this system lies in its use of a single triboelectric device—the iSD Sensor—to generate distinguishable signal types from different mechanical stimuli. Specifically, under sustained pressure, the sensor produces a stable and continuous signal, whereas transient mechanical inputs induce high-sensitivity, rapidly varying signals. These pressure-dependent responses are processed via a custom-designed DC/AC signal conversion module. The resulting output—comprising a continuous signal corresponding to static pressure and a transient, high-resolution signal corresponding to dynamic input—is then used to control various robotic actions such as grip adjustment, trigger gestures, and fine manipulation. This dual-mode signal architecture, derived from a single sensor and a unified working principle, significantly simplifies the system design while enhancing control precision and adaptability. It offers a scalable solution for self-powered human–machine interfaces, next-generation prosthetics, and intelligent assistive robotics. As shown in Fig. 1(b), the device adopts a multilayer sandwich structure, comprising a top and bottom acrylic substrate for mechanical support, an intermediate sponge interlayer for contact separation regulation, and an elastic conductive sponge layer serving as both a triboelectric component and soft electrode. An ePTFE film is positioned opposite the conductive sponge, forming an effective triboelectric pair. Aluminum foil is integrated beneath the conductive sponge to enhance charge collection and signal transmission. Fig. 1(c) presents a photograph of the fabricated iSD Sensor with a compact footprint of 4 cm × 4 cm, highlighting its potential for scalable integration. Fig. 1(d) shows the individual triboelectric materials: the conductive sponge offers excellent compressibility and conductivity, while the ePTFE film provides high electronegativity and a porous microstructure, both critical for effective pressure sensing. As shown in **Supplementary Note 1**, the ePTFE film features a flexible and porous multilayered microstructure, which increases the effective contact area and promotes charge retention. This structure is advantageous for triboelectric enhancement, enabling efficient signal generation under mechanical loading. Additionally, the elastic conductive sponge demonstrates good surface conductivity on both sides and integrates easily with the sensor architecture. Fig. 1(e) displays the custom-designed voltage multiplying circuit (VMC), which facilitates DC/AC signal separation and conditioning. Fig. 1(f) provides SEM image of the conductive sponge, revealing its highly porous internal architecture, which contributes to mechanical compliance and enhanced contact surface area. Fig. 1(g) illustrates the charge excitation strategy employed to enhance the signal resolution of the iSD Sensor. By applying an external pre-conditioning voltage using VMC, the surface charge density is increased, leading to improved signal differentiation under subtle pressure variations. This approach effectively boosts the iSD Sensor's capability to resolve fine pressure gradients in static sensing scenarios. Fig. 1(h) illustrates that the 3D gradient structural design of the conductive sponge introduces a built-in separation distance, which facilitates progressive contact formation under pressure. This design strategy significantly improves the sensing sensitivity of the iSD Sensor, particularly under low-pressure conditions.

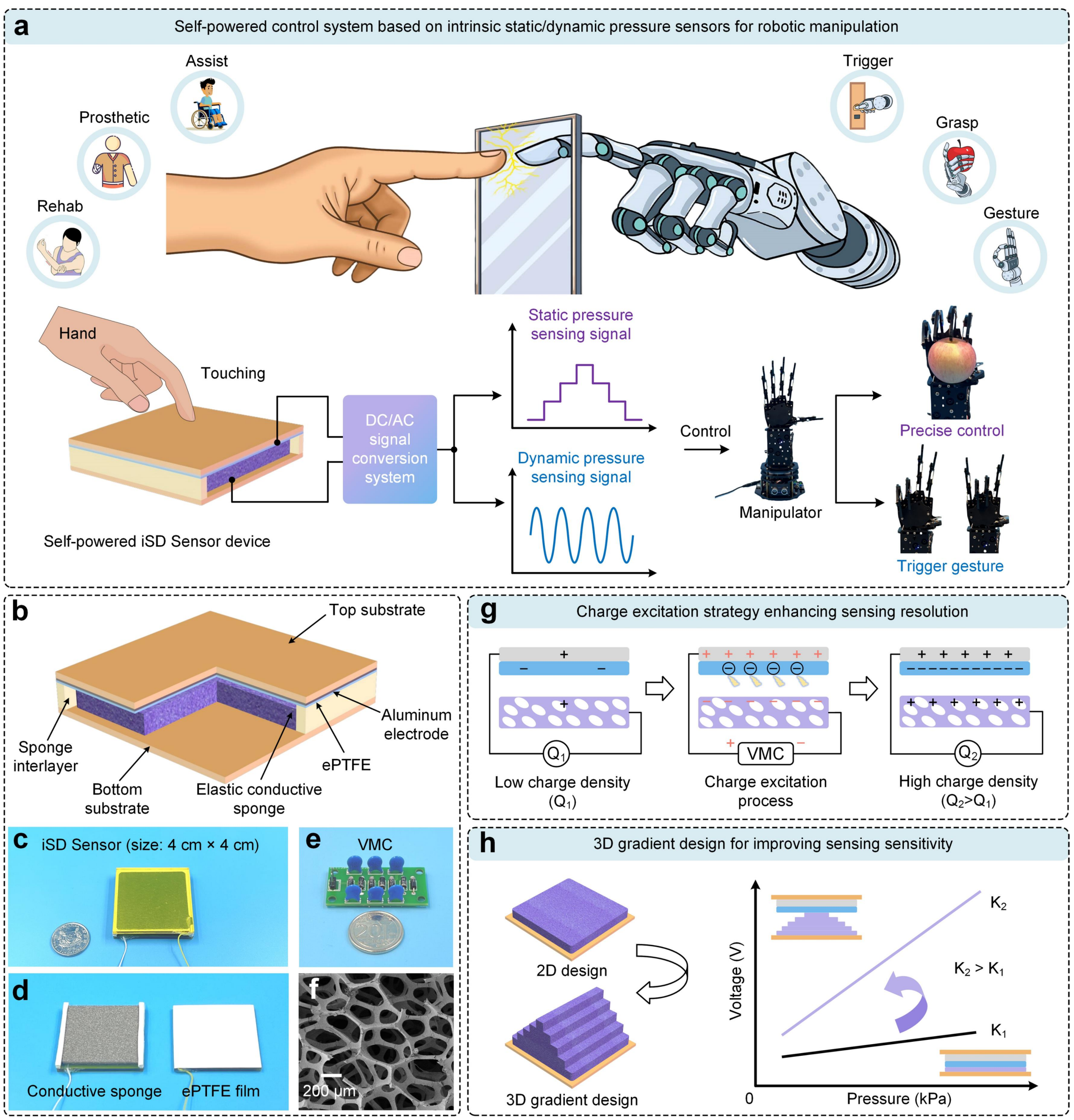

**Figure 1 | Self-powered control system enabled by intrinsic static/dynamic pressure sensing.** (a) Schematic illustration of the self-powered control concept based on the iSD Sensor, capable of generating distinguishable signals under static and dynamic mechanical stimuli. (b) Exploded structural diagram of the iSD Sensor, showing the multilayer configuration with acrylic substrates, elastic conductive sponge, ePTFE film, sponge interlayer, and aluminum electrode. (c) Photograph of the fabricated iSD Sensor (5 cm × 5 cm). (d) Optical images of the triboelectric materials: conductive sponge and ePTFE film. (e) Custom-designed VMC for signal processing and mode separation. (f) SEM image of the conductive sponge showing its porous structure. (g) Charge excitation (CE) strategy used to enhance sensing resolution by increasing surface charge density. (h) 3D gradient structural design of the conductive sponge introduces a built-in separation distance, enabling progressive contact and improving sensitivity compared to planar designs.

### 3.2. Boosting sensing resolution of iSD Sensor via charge excitation method.

Fig. 2(a) illustrates the measurement setup used for evaluating the static and dynamic pressure sensing performance of the iSD

Sensor. An exciter system applies controlled mechanical force to the iSD Sensor, which then generates pressure-dependent electrical signals. A custom-designed VMC circuit is employed to excite the transferred charge density in the iSD Sensor, leading to improved signal output and enhanced sensing performance. The resulting DC (for static pressure) and AC (for dynamic pressure) outputs are recorded by an electrometer and visualized in real time using an oscilloscope. This setup allows simultaneous monitoring of both signal types, facilitating comprehensive characterization of dual-mode sensing behavior. The static and dynamic pressure sensing performance of the iSD Sensor was evaluated using a custom-built excitation and measurement system (**Supplementary Note 2**). As shown in Fig. 2(b), two excitation modes are implemented: positive charge excitation (PCE) and reverse charge excitation (RCE). In the PCE mode, a positive bias is applied to induce additional negative charges on the ePTFE surface, while the RCE mode applies a negative bias to further polarize the interface. Both methods result in an increased net charge density at the contact interface, strengthening the triboelectric potential during subsequent mechanical interaction. The effectiveness of this approach is validated in Fig. 2(c), where the accumulated charge is significantly increased after applying either PCE or RCE. This enhancement is further reflected in the dynamic voltage output, as shown in Fig. 2(d), where the signal amplitude increases from ~2.1 nC to ~39.3 nC—a ~18.7-fold improvement—following charge excitation. This substantial increase in output signal underlines the efficacy of the CE strategy in boosting the iSD Sensor's sensitivity and resolution, particularly for applications requiring fine pressure discrimination. To mitigate the limitations associated with real-time instability during charge excitation, the VMC module was removed after CE, allowing the iSD Sensor to operate in a stabilized, pre-excited mode for practical sensing applications (see **Supplementary Note 3**).

### 3.3. Mechanism and enhancement of static/dynamic pressure sensing of iSD Sensor.

Fig. 2(e) illustrates the working mechanism of static pressure sensing enabled by the iSD Sensor, which relies on charge retention and signal stability under sustained contact. Notably, static pressure sensing is achieved through microscale contact–separation at the triboelectric interface, without relying on mechanical elasticity from supporting structures. Fig. 2(e1) shows the moment when external pressure is applied, compressing the conductive sponge and ePTFE film into tight contact. This close interface ensures efficient charge transfer between the two triboelectric layers. Fig. 2(e2) depicts the onset of pressure release, during which a slight separation emerges between the two layers. This microscale decoupling preserves the previously generated surface charges and initiates the formation of a stable potential difference. Fig. 2(e3) represents the state after complete pressure removal. In this condition, contact between the two triboelectric films is sustained only by the device's own weight, without active mechanical compression. This minimal-contact state maintains a weak residual signal or returns to baseline, depending on the structural configuration. Fig. 2(e4) illustrates the subsequent reapplication of pressure, during which the interfacial contact is re-established and microscale contact is enhanced. This renewed interaction generates a detectable triboelectric signal, enabling repeatable static pressure sensing. Fig. 2(f) illustrates the working mechanism of dynamic pressure sensing in the iSD Sensor, which is driven by rapid contact–separation cycles at the triboelectric interface. In contrast to static sensing, this mode relies on the elastic deformation of supporting components to enable repeated mechanical motion. Importantly, a fixed separation gap is engineered between the two triboelectric layers under unloaded conditions, ensuring a distinct dynamic response upon impact. Fig. 2(f1) shows the fully compressed state under dynamic pressure input. The conductive sponge and ePTFE film are in complete contact, facilitating triboelectric charge transfer at the interface. Fig. 2(f2) depicts the initial stage of unloading, where the applied pressure begins to decrease and a separation gap starts to form between the two triboelectric layers. This dynamic decoupling induces charge redistribution and generates a transient sensing signal in the external circuit. Fig. 2(f3) represents the fully separated state after the pressure is completely released. The two triboelectric layers return to their preset maximum separation distance due to the elastic recovery of the supporting structure. At this moment, the electrostatic potential difference reaches its peak, resulting in the maximum sensing voltage in the external circuit. Fig. 2(f4) illustrates the reapplication of pressure, where the separation gap decreases and the layers re-engage in contact. This process initiates a new charge transfer cycle, enabling repeatable and high-sensitivity dynamic sensing. Fig. 2(g) shows the output under static pressure of iSD Sensor before and after CE. Before CE, the signal amplitude remains below 1.5 V, indicating limited surface charge accumulation. After CE, the voltage output increases sharply to ~38.2 V, corresponding to a ~25.4-fold enhancement. This significant improvement confirms the effectiveness of CE in boosting charge density and enhancing the resolution of static pressure sensing. Fig. 2(h) presents the dynamic pressure sensing response of iSD Sensor before and after CE. Without CE, the transient voltage peaks are weak and difficult to distinguish. Following CE treatment, the output signal reaches ~163.6 V, representing a ~15.2-fold increase in sensitivity. The amplified peak signals ensure high signal-to-noise ratio and accurate detection of rapid mechanical events. Together, these results demonstrate that the CE strategy substantially enhances both static and dynamic sensing performance, enabling the iSD Sensor to deliver robust, high-resolution output signals across a wide range of mechanical stimuli.

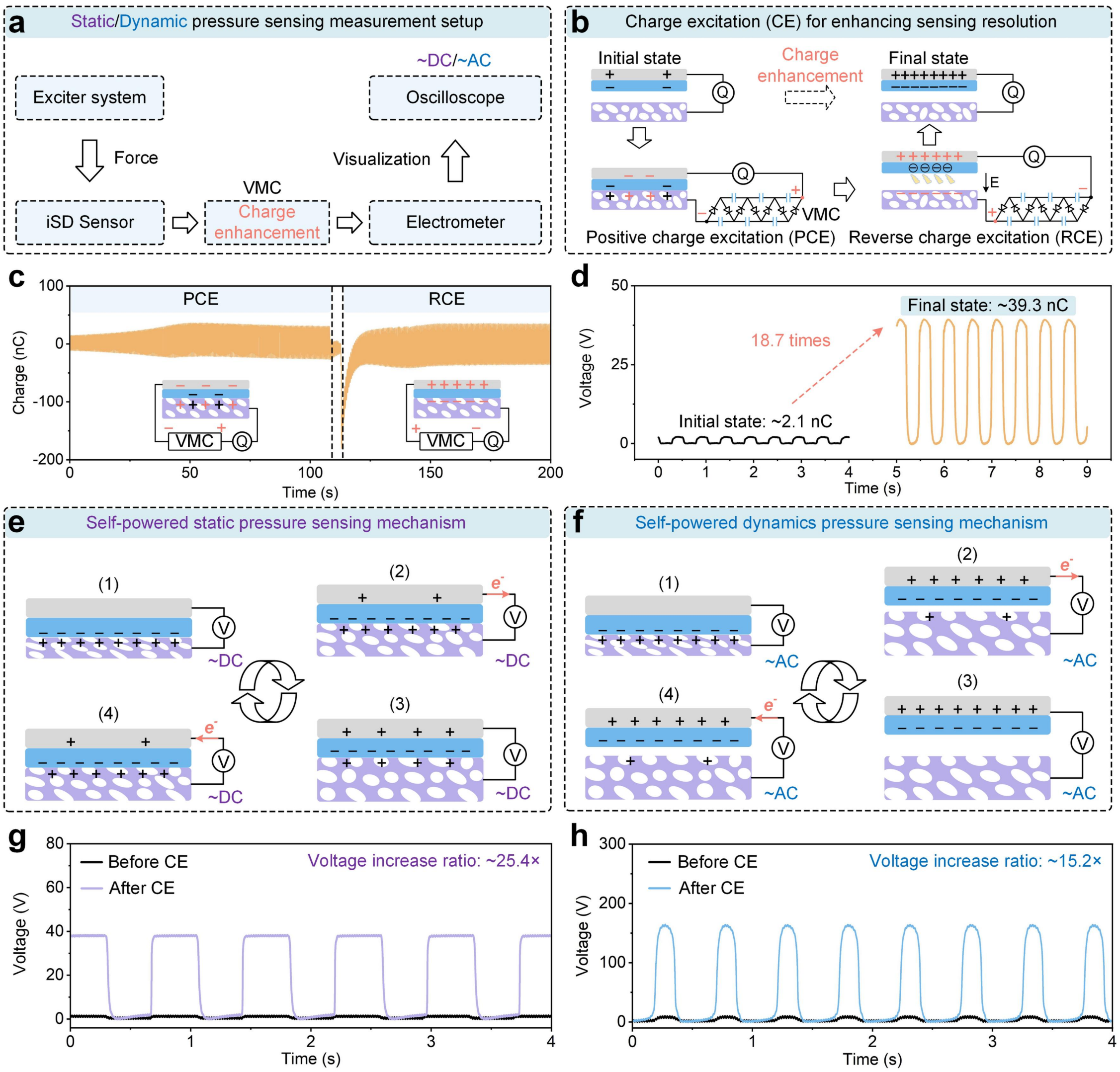

**Figure 2 | Charge excitation strategy and static/dynamic sensing mechanism of the iSD Sensor.** (a) Schematic of the measurement setup for static and dynamic pressure sensing. (b) Illustration of the CE strategy to enhance surface charge density on the ePTFE layer. (c) Accumulated charge output during PCE and RCE processes, demonstrating effective surface charge enhancement. (d) Dynamic voltage response before and after CE, showing a ~18.7-fold increase in peak signal amplitude. (e1-e4) Working mechanism of static pressure sensing of iSD Sensor. (f1-f4) Working mechanism of dynamic pressure sensing of iSD

Sensor. (g) Static voltage output before and after CE, showing a ~25.4× increase in signal amplitude. (h) Dynamic voltage output before and after CE, showing a ~15.2× increase in signal amplitude.

### 3.4. Performance evaluation of iSD Sensor for static and dynamic pressure sensing.

Fig. 3(a) shows the measurement setup for static pressure sensing. A square-wave mechanical excitation is applied using a programmable exciter, while pressure and voltage signals are recorded simultaneously via a digital pressure gauge and voltmeter. Fig. 3(b) displays the iSD Sensor's voltage output as a function of applied static pressure. The response curve reveals three distinct sensing regions: a high-sensitivity region (0.6–9.8 kPa) with a slope of 2.6 V·kPa$^{-1}$, where micro-contact dominates the charge generation; a transition region (12.2–24.5 kPa) with moderate sensitivity (0.7 V·kPa$^{-1}$), indicating progressive surface engagement; and a low-sensitivity saturation region (>27.5 kPa) with reduced slope (0.2 V·kPa$^{-1}$), where full contact limits further voltage growth. This multi-region response enables both low-force detection and large-force tolerance. Fig. 3(c) presents real-time voltage signals under cyclic loading at different pressure levels, showing stable, repeatable outputs with distinguishable amplitudes proportional to applied pressure. Fig. 3(d) characterizes the temporal response, demonstrating a fast response time of ~83 ms and recovery time of ~43 ms, suitable for real-time tactile monitoring. Fig. 3(e) illustrates the iSD Sensor's capacity for discrete load recognition. Stepwise application and removal of calibrated weights (50–200 g) result in voltage plateaus corresponding to each load increment. The inset shows the sensor setup with the suspended weight module. These results collectively confirm the iSD Sensor's capability for high-resolution, broad-range static pressure sensing with fast and reliable signal response. To further demonstrate the long-term reliability of the iSD Sensor, its dynamic output was evaluated after ambient storage without re-excitation. As shown in **Supplementary Note 4**, the iSD Sensor retained stable signal amplitude and periodicity over three weeks, benefiting from its intrinsic static sensing configuration and sealed construction that mitigates environmental charge dissipation.

To interpret the static pressure sensing behavior of the iSD Sensor, we established an electrostatic model treating the sensor as a deformable parallel-plate capacitor. Under increasing external pressure, the contact area between the conductive sponge and the ePTFE layer expands, and the interfacial distance decreases, resulting in a voltage rise. The $V_{OC}$ is governed by:

$$V_{OC} = \frac{Q}{C} = \frac{Qd}{\varepsilon_0 \varepsilon_r A} \quad (1)$$

where $Q$ is the transferred charge, $d$ is the ePTFE thickness (compressible), $A$ is the effective contact area, and $\varepsilon_0$, $\varepsilon_r$ represent the permittivity of air and ePTFE layer. Both $A$ and $d$ vary with pressure. Substituting their pressure-dependent expressions yields:

$$V_{static}(P) = \frac{Q(d-\beta P)}{\varepsilon_0 \varepsilon_r (A_0+\alpha P)} \quad (2)$$

where $\alpha$ and $\beta$ are empirical coefficients describing area expansion and thickness compression, respectively. This relation explains the nonlinear voltage-pressure response observed experimentally (see Fig. 3(b)), and is fully derived in **Supplementary Note 5**.

As pressure increases, the iSD Sensor exhibits three distinct sensitivity regions. In the low-pressure regime (0.6–9.8 kPa), the rapid increase in contact area and sharp reduction in dielectric spacing jointly contribute to a high sensitivity of 2.6 V·kPa$^{-1}$. This corresponds to the steep slope region in Equation (2), where both numerator and denominator vary significantly with pressure. In the transition region (12.2–24.5 kPa), the rate of area expansion slows and compression becomes less pronounced, resulting in a moderate sensitivity of 0.7 V·kPa$^{-1}$. Finally, in the saturation regime (>27.5 kPa), the structural deformation approaches its limit, and the sensitivity declines to 0.2 V·kPa$^{-1}$, consistent with the flattening behavior predicted when both $A(P)$ and $d(P)$ reach asymptotic bounds (see Equation (5-9) and its derivative in **Supplementary Note 5**).

Fig. 3(f) shows the dynamic measurement setup, in which sinusoidal mechanical excitation is applied using a programmable exciter. The iSD Sensor output is recorded in real time using an oscilloscope for AC signal analysis. To determine the optimal preset separation for dynamic pressure sensing, the iSD Sensor was evaluated under increasing gap distances from 1 mm to 8 mm. As shown in **Supplementary Note 6**, the output signals ($V_{OC}$, short-circuit current ($I_{SC}$), and transfer charge ($Q_{SC}$)) increased with separation up to ~4 mm, beyond which no significant improvement was observed. This plateau suggests that excessive separation yields diminishing returns. A 1 mm gap was selected for final device implementation, as it enhances output resolution while maintaining structural compactness. In contrast, larger gaps (e.g., 2–3 mm) would require bulkier support components, reducing sensitivity to low-pressure stimuli and limiting practical integration. Fig. 3(g) presents the output voltage as a function of dynamic pressure amplitude. Similar to the static case, the dynamic response can be divided into three regions: a high-sensitivity region (0–5.5 kPa) with a peak slope of 9.8 V kPa$^{-1}$, a transition region (7.3–12.2 kPa) with a sensitivity of 2.3 V kPa$^{-1}$, and a saturation region (>15.3 kPa) with reduced sensitivity (~0.1 V·kPa$^{-1}$). This layered sensitivity ensures accurate perception of both small and large dynamic inputs. Fig. 3(h) displays dynamic voltage outputs under periodic loading at different pressures. The signal amplitude increases with pressure, showing clear differentiation across levels from 1.2 to 7.4 kPa.

The dynamic response of the iSD Sensor originates from a periodic contact–separation process between the triboelectric layers, as depicted in **Supplementary Note 7**. In this mode, the surface charge generated upon contact is released as a transient voltage signal when the layers separate, forming an alternating output governed by electrostatic modulation.

The output voltage $V_{dynamic}$ can be theoretically described as:

$$V_{dynamic}(P) = \frac{\sigma(P)(x+d)}{\varepsilon_0 \varepsilon_{eff}} \tag{3}$$

where $\sigma(P)$ is the pressure-dependent surface charge density, $x$ is the fixed separation gap (1 mm), and $d$ is the dielectric thickness of the ePTFE layer.

To capture the experimentally observed nonlinear behavior, the pressure–voltage response is fitted with an exponential model:

$$V_{dynamic}(P) = V_{max}(1 - e^{-kP}) \tag{4}$$

where $V_{max}$ represents the saturation voltage and $k$ is a constant related to material compliance.

The sensitivity $S$ can be derived as the first-order derivative:

$$S_{dynamic} = \frac{dV_{dynamic}}{dP} = V_{max} \cdot k \cdot e^{-kp} \tag{5}$$

This formulation indicates that sensitivity is highest at low pressure and decays exponentially as pressure increases. The dynamic pressure response of the iSD Sensor can be clearly divided into three sensitivity regions based on the applied pressure. In the low-pressure region (0–5.5 kPa), the sensor exhibits high sensitivity (9.8 V·kPa$^{-1}$), attributed to the rapid increase in effective contact area and charge density during the initial compression cycle. As pressure enters the transition region (7.3–12.2 kPa), the rate of increase in surface contact begins to saturate, resulting in a moderate sensitivity of 2.3 V·kPa$^{-1}$. In the high-pressure or saturation region (>15.3 kPa), the triboelectric layers are nearly fully compressed, and further increases in pressure yield minimal changes in contact conditions or charge output, reducing the sensitivity to ~0.1 V·kPa$^{-1}$. This three-region behaviour is well explained by the dynamic sensing model derived in **Supplementary Note 7**, and matches the exponential voltage–pressure trend observed in experimental results.

Fig. 3(i) shows the frequency response of the sensor across a range of input frequencies (2–6 Hz). The sensor maintains stable waveform characteristics, demonstrating robust performance under various dynamic conditions relevant to human motion and robotic interaction. The iSD Sensor demonstrates consistent dynamic sensing performance across a range of excitation frequencies. As shown in **Supplementary Note 8**, the $I_{SC}$ increases with frequency due to enhanced charge transport under faster mechanical cycles, while the $Q_{SC}$ remains nearly unchanged, indicating stable output behavior and robust signal integrity under varying dynamic conditions. Fig. 3(j) evaluates long-term durability. After 30,000 continuous loading cycles, the output voltage remains stable with negligible degradation, confirming the mechanical and electrical stability of the device under cyclic dynamic operation. These results highlight the iSD Sensor's excellent dynamic pressure sensing capability, with high sensitivity, broad frequency adaptability, and outstanding durability for practical and wearable applications. **Supplementary Video 1** demonstrates the dynamic contact behavior of a water droplet on the ePTFE film surface. As the droplet advances and recedes under external force, the contact line remains stable with minimal deformation, indicating low contact angle hysteresis and excellent hydrophobic stability. This consistent wetting behavior ensures reliable surface performance under dynamic conditions. The iSD Sensor exhibits excellent environmental stability, maintaining consistent static and dynamic outputs under varying humidity (15–85%) and temperature (20–53 °C) conditions (**Supplementary Note 9**). To further demonstrate its self-powered capability, the iSD Sensor was integrated into a bridge-rectifier charging circuit and evaluated under different capacitances and actuation frequencies (**Supplementary Note 10**). Results show that higher frequencies and lower capacitance values lead to faster voltage accumulation, confirming the sensor's ability to serve not only as a dual-mode pressure detector but also as a micro-energy harvester for powering low-power electronics.

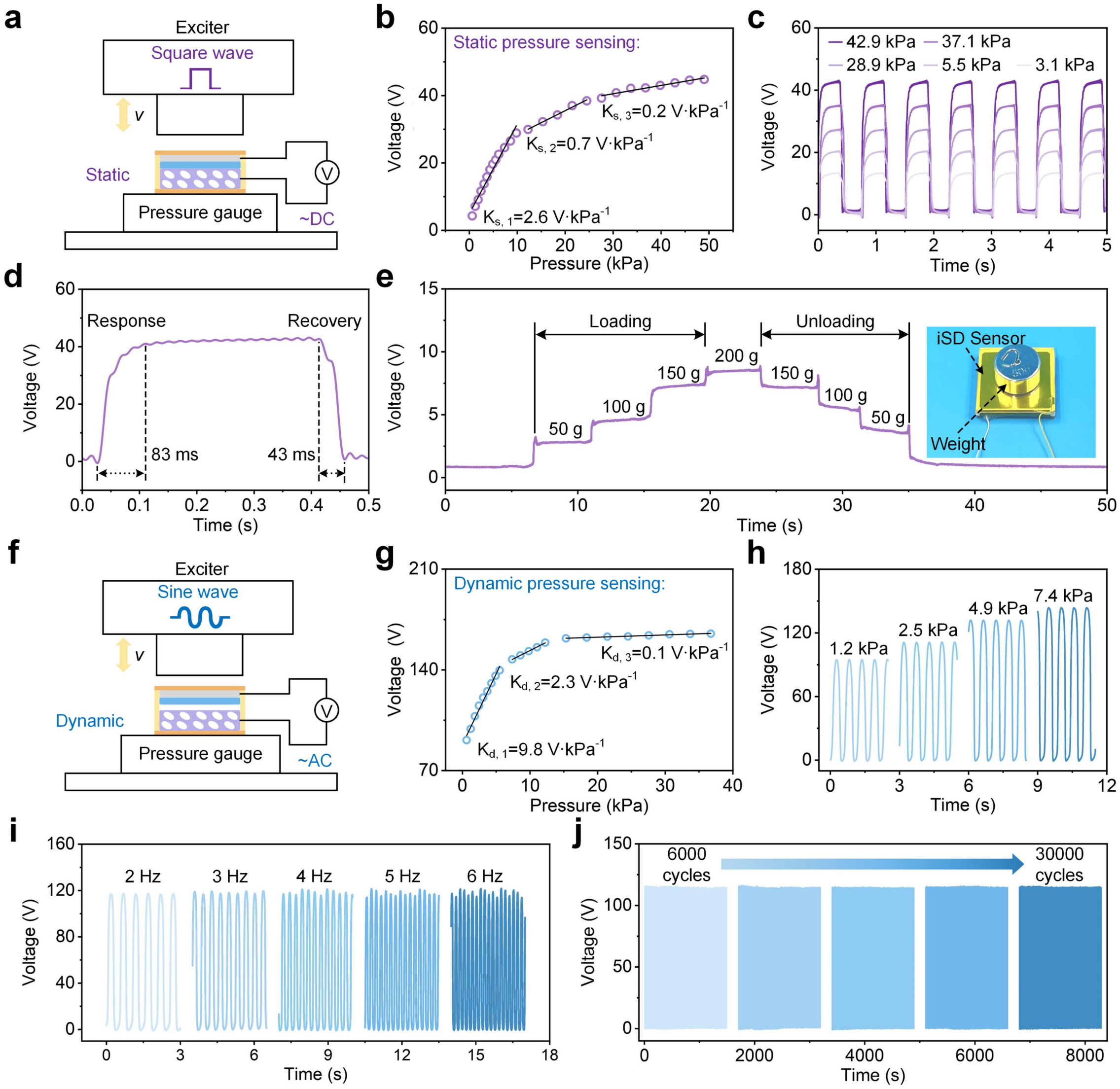

**Figure 3 | Comprehensive performance characterization of the iSD Sensor under static and dynamic pressure stimuli.** (a) Schematic diagram of the static pressure sensing setup using square-wave mechanical excitation. (b) Voltage output as a function of applied static pressure, exhibiting three sensing regions with different sensitivities: high (0.6–9.8 kPa, 2.6 V·kPa$^{-1}$), intermediate (12.2–24.5 kPa, 0.7 V kPa$^{-1}$), and saturation (>27.5 kPa, 0.2 V kPa$^{-1}$). (c) Stable and repeatable voltage responses under cyclic static loading at various pressure levels. (d) Transient response curve showing rapid response (~83 ms) and recovery (~43 ms) times under step pressure input. (e) Stepwise voltage outputs in response to sequential loading/unloading of discrete weights (50–200 g), demonstrating accurate force discrimination. Inset: photo of the experimental setup with calibrated weights. (f) Schematic of the dynamic pressure sensing setup using sinusoidal excitation. (g)

Voltage–pressure relationship under dynamic conditions, also showing three sensitivity regions: high (0.6–5.5 kPa, 9.8 V $kPa^{-1}$), intermediate (7.3–12.2 kPa, 2.3 V $kPa^{-1}$), and saturation (>15.3 kPa, 0.1 V $kPa^{-1}$). (h) Distinct voltage outputs under different dynamic pressure levels from 1.2 to 7.4 kPa. (i) Frequency response of the sensor under periodic excitations from 2 to 6 Hz, with consistent waveform stability. (j) Long-term durability test showing consistent voltage outputs over 30,000 mechanical loading cycles.

### 3.5. Performance enhancement of static and dynamic pressure sensing via a 3D gradient architecture.

Fig. 4(a) shows a schematic of the 3D gradient structure composed of five stacked layers of conductive sponge with gradually increasing contact surface area from top to bottom. This layered configuration introduces a built-in separation distance and enables progressive contact under pressure, effectively modulating the contact area and pressure distribution during compression. Fig. 4(b) depicts the sensing mechanism associated with this structure. In the initial state, the triboelectric layers are separated by the gradient profile, with minimal contact and low output. Upon compression, the layers engage progressively from top to bottom, allowing for more effective charge transfer and a stronger signal output under low-pressure stimuli. Fig. 4(c) presents a photograph of the fabricated 3D gradient sponge structure, confirming its stepwise configuration and compact size. The 3D gradient structure of the iSD Sensor was assembled by stacking conductive sponge layers with increasing contact area, as detailed in **Supplementary Note 11**. The design offers tunable mechanical response, improved signal resolution, and enhanced sensitivity, particularly in the low-pressure regime, making it well-suited for high-resolution tactile sensing applications. To validate the electrical reliability of the 3D gradient design, a conductivity test was conducted on the stacked conductive sponge layers. As shown in **Supplementary Video 2**, the structure maintains continuous electrical conductivity across all layers, confirming that no interfacial resistance or signal interruption occurs due to stacking. This ensures reliable signal transmission in pressure sensing applications, even under multilayer configurations. Fig. 4(d) shows the experimental setup for static pressure sensing, using square-wave mechanical excitation. The pressure is monitored via a digital gauge while the voltage output is recorded. Fig. 4(e) illustrates the effect of gradient layer number on the output voltage under a constant applied pressure of 1.8 kPa. As the number of conductive sponge layers increases from 2 to 5, the output voltage rises significantly. This enhancement is attributed to the larger initial separation gap introduced by the 3D gradient structure, which facilitates greater deformation and contact area upon compression. As a result, more surface charge is generated at the triboelectric interface, leading to stronger electrical signals even under the same applied pressure. Fig. 4(f) shows the voltage–pressure response curve, which exhibits two distinct sensitivity regions. In the low-pressure regime (0.006–1.8 kPa), the sensor achieves a high sensitivity of 34.7 V·$kPa^{-1}$, while in the moderate range (2.1–3.6 kPa), the sensitivity is 15.8 V·$kPa^{-1}$. These values significantly surpass those of conventional planar structures, confirming the performance benefits of the 3D gradient design. Fig. 4(g) provides stable and distinguishable output waveforms under repeated loading at pressures from 0.6 to 3.7 kPa, demonstrating excellent resolution and signal reproducibility. Fig. 4(h) further highlights the iSD Sensor's ultralow pressure detection capability, with clear voltage responses measurable down to a minimum pressure of 6.13 Pa, validating its potential for high-resolution tactile sensing in lightweight or subtle force scenarios. Fig. 4(i) shows the dynamic testing setup using sinusoidal mechanical excitation. Pressure is modulated through a programmable exciter, and the voltage signal is recorded via an oscilloscope for AC signal analysis. Fig. 4(j) depicts the voltage–pressure response of the sensor under dynamic loading. A high sensitivity of 48.4 V $kPa^{-1}$ is achieved in the low-pressure regime (0.06–1.8 kPa), and 9.1 V $kPa^{-1}$ in the intermediate range (2.1–3.6 kPa). These values indicate a significant enhancement compared to non-gradient configurations, resulting from the progressive contact and increased deformability of the 3D structure. Fig. 4(k) presents real-time dynamic output signals under various pressures ranging from 0.5 to 3.7 kPa. The iSD Sensor produces stable, periodic waveforms with clearly distinguishable amplitudes corresponding to different pressure levels. The signal-to-pressure correlation further confirms the high resolution and repeatability of the dynamic sensing mode. These results validate the iSD Sensor's capability to detect subtle and fast-changing pressure variations with high fidelity, making it a promising platform for real-time human–machine interface applications, soft robotics, and wearable electronics.

To elucidate the sensitivity enhancement enabled by the 3D gradient architecture, we attribute the performance improvement to three key mechanisms: progressive contact, stress redistribution, and enhanced charge generation. Unlike planar structures where full contact occurs rapidly upon compression, the stacked gradient layers in the iSD Sensor introduce a built-in separation gap and enable sequential engagement from top to bottom under increasing pressure. This progressive contact formation effectively prolongs the active sensing duration and facilitates more gradual, high-resolution signal variation under low-pressure stimuli. Moreover, the layered structure redistributes applied stress more uniformly, preventing early saturation and allowing for continued deformation and charge transfer even at relatively small forces. Each additional layer contributes incrementally to the total contact area, resulting in cumulative charge accumulation and amplified output signals. These combined effects significantly boost the sensitivity in both static and dynamic sensing modes, particularly in the low-pressure regime, as evidenced by the observed enhancement from 2.6 to $34.7\ V\cdot kPa^{-1}$ and from 9.8 to $48.4\ V\cdot kPa^{-1}$. Thus, the 3D gradient design provides an effective structural pathway to achieve ultra-sensitive, wide-range tactile perception.

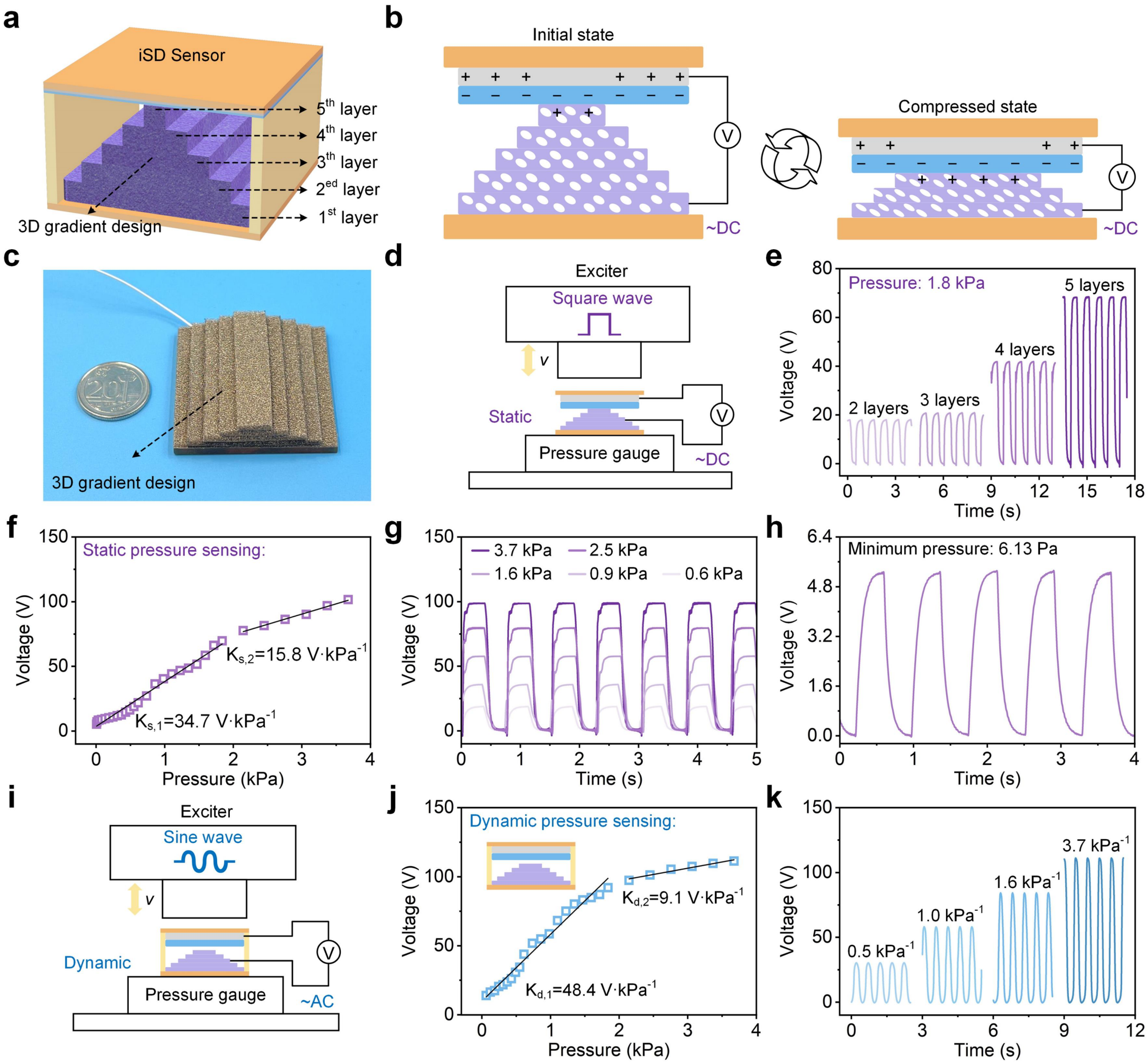

**Figure 4 | Enhanced static and dynamic pressure sensing enabled by a 3D gradient structural design.** (a) Schematic of the 3D gradient structure consisting of five stacked conductive sponge layers with gradually increasing surface area from top to bottom, introducing a built-in separation gap and enabling progressive contact under compression. (b) Working principle of the gradient structure. Initial separation ensures low baseline signal, while progressive contact under loading enhances charge generation. (c) Photograph of the fabricated 3D gradient

conductive sponge with a stepwise configuration. (d) Experimental setup for static pressure sensing of iSD Sensor. (e) Output voltage comparison under a constant pressure of 1.8 kPa for sensors with different gradient layer numbers. Increased separation distance from additional layers leads to higher signal output. (f) Voltage–pressure response curve under static pressure, showing high sensitivity of 34.7 V kPa$^{-1}$ (0.006–1.8 kPa) and 15.8 V kPa$^{-1}$ (2.1–3.6 kPa). (g) Real-time voltage outputs under repeated static loading at pressures from 0.6 to 3.7 kPa, exhibiting clear signal distinction and stability. (h) Ultrahigh sensitivity demonstrated by successful detection of ultralow pressure down to 6.13 Pa. (i) Experimental setup for dynamic pressure sensing using sinusoidal excitation. (j) Voltage–pressure response under dynamic loading, with sensitivity of 48.4 V kPa$^{-1}$ (0.06–1.8 kPa) and 9.1 V kPa$^{-1}$ (2.1–3.6 kPa). (k) Dynamic output waveforms under pressures ranging from 0.5 to 3.7 kPa, showing stable, distinguishable signal amplitudes.

To further benchmark the performance of the iSD Sensor, a comparative analysis of previously reported TENG-based pressure sensors focusing on low-pressure sensitivity is presented in Table 1. The table summarizes representative works covering both static and dynamic modes under low-pressure conditions, along with their respective sensing mechanisms, measurement setups, and reported sensitivities. Most existing studies employed independent measurement schemes with separate equipment for static or dynamic signal acquisition, often resulting in inconsistent calibration conditions. Furthermore, most of the devices exhibit limited capability in either static or dynamic sensing alone. Notably, only a few designs report static sensitivities exceeding 10 V·kPa$^{-1}$ at low pressures, and even fewer demonstrate reliable detection in both modes simultaneously. Our work demonstrates simultaneous, high-resolution sensing in both static and dynamic modes using a single integrated sensor and a unified DC/AC measurement scheme. The iSD Sensor achieves a static sensitivity of 34.7 V·kPa$^{-1}$ (<1.8 kPa) and a dynamic sensitivity of 48.4 V·kPa$^{-1}$ (<1.8 kPa). These results confirm the iSD Sensor's superior performance in low-pressure regimes and its suitability for high-fidelity tactile robotics, where both sustained and transient signals are critical for intelligent actuation. Compared with previously reported TENG sensors (Table 1), the iSD Sensor uniquely achieves simultaneous high-resolution static and dynamic pressure sensing, as further demonstrated by the detailed output characteristics in Fig. 3 and Fig. 4.

Table 1. Comparison of low-pressure sensitivity in TENG-based static and dynamic pressure sensors from previous studies.

| Reference | Mechanism | Measurement Method | | Pressure Sensing Sensitivity | |
|---|---|---|---|---|---|
| | | Equipment | Usage Mode | Static | Dynamic |
| Ref. [43] | Triboelectric Capacitive | Oscilloscope | Single Device Measurement | 12.062V·kPa$^{-1}$ (<3 kPa) | 0.025 V·kPa$^{-1}$ (<3 kPa) |
| Ref. [49] | Triboelectric +MOSFET | Electrometer MCU-MSP40 | Independent Measurement | None, (<0.24 kPa) | 53.7 V·kPa$^{-1}$ (<0.7 kPa) |
| Ref. [50] | Piezoresistive Triboelectric | Oscilloscope Source Meter | Independent Measurement | 1.69 Kpa$^{-1}$ (<12 kPa) | 0.58 V·kPa$^{-1}$(<~100 kPa) |
| Ref. [44] | Piezoelectric Triboelectric Capacitive | Electrometer LCR meter Oscilloscope | Independent Measurement | 0.457 kPa$^{-1}$ (97-134 kPa) | None |
| Ref. [45] | Piezoresistive Triboelectric | Electrometer Source Meter | Independent Measurement | 126135.9 kPa$^{-1}$ (<1.2 kPa) | None |
| Ref. [48] | Triboelectric | Electrometer Oscilloscope | Simultaneous Dual-Mode (DC & AC) | 1.48 V·kPa$^{-1}$ (<6.3 kPa) | 10.09 V·kPa$^{-1}$ (<5.6 kPa) |
| **This work** | **Triboelectric** | **Electrometer Oscilloscope** | **Simultaneous Dual-Mode (DC & AC)** | **34.7 V·kPa$^{-1}$ (<1.8 kPa)** | **48.4 V kPa-1 (<1.8 kPa)** |

### 3.6. Demonstration of iSD Sensor for robotic control and assistive interaction.

Fig. 5(a) shows the experimental setup of the self-powered sensing–control system. The iSD Sensor is connected to a Keithley 6517B electrometer for signal measurement. The analog output from the electrometer is transmitted to a custom-designed data acquisition module, where the signal is conditioned and processed. The resulting control signals are wirelessly transmitted via an RF module to the robotic manipulator's control terminal, enabling real-time, closed-loop actuation in response to tactile inputs. Fig. 5(b) illustrates the signal conditioning and control architecture of the sensing-actuation system. The iSD Sensor is connected to a signal conditioning circuit comprising a capacitor (C) and resistor (R), which are used to modulate the amplitude and pulse width of the sensing signals. The processed signals are measured by a Keithley 6517B electrometer and subsequently transmitted to the signal conversion and data acquisition modules. After digital processing and wireless transmission, the control commands are delivered to the robotic manipulator, forming a closed-loop feedback system for real-time motion control. Fig. 5(c) investigates the effect of varying capacitance values in the signal conditioning circuit on the output voltage of the iSD Sensor. As the capacitance increases, the signal amplitude is effectively modulated, enabling control over the voltage response under identical pressure input. This result confirms that the capacitor serves as a tunable element to adjust the signal strength and match the downstream processing requirements, providing a flexible mechanism for optimizing sensor output in different application scenarios. Fig. 5(d) illustrates the impact of varying series resistance on the output waveform of the iSD Sensor. As the resistance increases from 50 MΩ to 500 MΩ, the pulse width of the voltage signal becomes progressively broader. This tunable response enables precise alignment with the temporal requirements of the robotic trigger mechanism, allowing optimal matching between the sensing signal and the manipulator's actuation threshold. To verify dual-mode signal characteristics under realistic inputs, we performed manual tests using finger pressure to stimulate the iSD Sensor. As shown in **Supplementary Note 12**, static pressing generated stable, continuous voltage outputs, while dynamic tapping induced distinct bipolar spikes. These differentiated responses support the sensor's application in dual-mode robotic control, where static inputs enable proportional actuation and dynamic inputs trigger rapid responses. Fig. 5(e) shows the bending behavior of the robotic fingers under varying pressure inputs sensed by the iSD Sensor. With increasing pressure, the output voltage gradually rises from -0.152 V to 4.014 V, resulting in proportionally enhanced finger bending amplitudes. This demonstrates the system's ability to convert analog pressure signals into graded motion outputs, enabling smooth and precise actuation. The entire dynamic process, including continuous pressure application and finger bending, is captured in **Supplementary Video 3**, further validating the system's real-time responsiveness and resolution in self-powered closed-loop control. Fig. 5(f) demonstrates the robotic manipulator's ability to perform accurate object grasping under the control of the iSD Sensor. The system successfully recognizes and responds to different tactile pressure inputs, enabling stable and adaptive grasping of diverse items, including a banana, a rubber rod, and a beverage bottle. These results highlight the sensor's high sensitivity and the closed-loop system's capacity to drive precise manipulation of objects with varying sizes and compliance. The detailed grasping processes are shown in **Supplementary Video 4** (banana), **Supplementary Video 5** (rubber rod), and **Supplementary Video 6** (beverage bottle), confirming real-time self-powered pressure sensing and control performance in practical handling scenarios. To further demonstrate the application potential of the iSD Sensor in inclusive human-machine interaction, we developed a sign language communication system for individuals with disabilities. As shown in Fig. 5(g) and **Supplementary Video 7**, a disabled user can transmit tactile information by touching the iSD Sensor, which then wirelessly triggers predefined robotic hand gestures for sign language output. Fig. 5(h) illustrates the successful execution of sequential gestures ('one', 'two', and 'three') in response to distinct pressure inputs. This scenario highlights the practical value of the iSD Sensor in enabling intuitive, pressure-based robotic control for assistive communication.

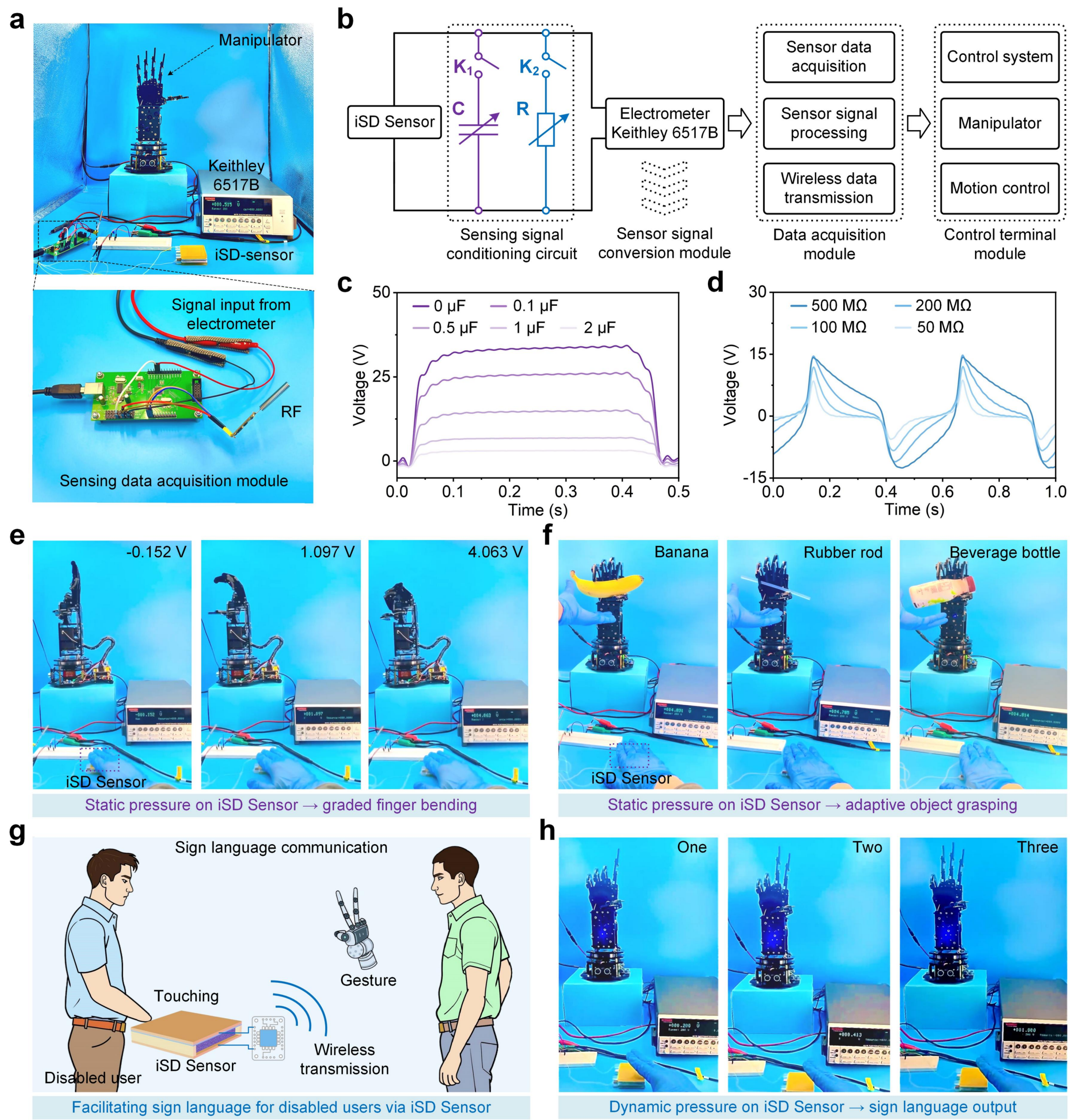

**Figure 5 | Application of the iSD Sensor for robotic control and human-machine interaction.** (a) Experimental setup showing the iSD Sensor connected to an electrometer and a manipulator for pressure-driven robotic control. (b) Schematic of the signal conditioning and wireless transmission system, including capacitance (C) and resistance (R) components for signal tuning. (c) Modulation of output voltage amplitude by varying the parallel capacitance (0–2 μF), enabling control of robotic actuation intensity. (d) Adjustment of output signal pulse width via series resistance (50–500 MΩ) to optimize timing for robotic trigger response. (e) Robotic finger bending under different pressure levels, demonstrating pressure-dependent actuation with corresponding output voltages. (f) Accurate grasping of various objects (banana, rubber rod, beverage bottle) triggered by tactile signals from the iSD Sensor. (g) Conceptual illustration of sign language communication: a disabled user interacts with the iSD Sensor to wirelessly trigger robotic hand gestures. (h) Robotic gesture communication based on pressure-triggered commands, displaying distinct hand signs ("One," "Two," "Three") corresponding to different input signals.

## 4. Conclusion

In this work, we develop a self-powered intrinsic static–dynamic (iSD) pressure sensor capable of high-resolution, decoupled pressure sensing for intelligent control applications. A charge excitation (CE) strategy significantly enhances interfacial charge density, yielding 25.4-fold and 15.2-fold improvements in static and dynamic voltage outputs, respectively. The sensor exhibits multi-region sensitivity profiles: 2.6, 0.7, and 0.2 $V\cdot kPa^{-1}$ for static stimuli across 0.6–9.8, 12.2–24.5, and >27.5 $kPa^{-1}$; and 9.8, 2.3, and 0.1 $V\cdot kPa^{-1}$ for dynamic stimuli across 0–5.5, 7.3–12.2, and >15.3 kPa. It demonstrates fast response/recovery times (~83 ms / ~43 ms), high durability (>30,000 cycles), and stable performance under dynamic loading (2–6 Hz). To enhance low-pressure responsiveness, a 3D stacked gradient structure is introduced to facilitate progressive contact behavior, resulting in ultrahigh sensitivities of 34.7 $V\cdot kPa^{-1}$ (static) and 48.4 $V\cdot kPa^{-1}$ (dynamic) below 1.8 kPa, with a minimum detection limit of 6.13 Pa. Integrated into a wireless closed-loop system, the iSD Sensor enables dual-mode control: static signals regulate proportional output, while dynamic inputs activate fast triggering functions. Demonstrations including graded motion control, adaptive grasping, and pressure-triggered communication validate its real-time responsiveness and functional versatility. Overall, this work presents a unified, scalable pressure-sensing control platform that intrinsically separates and utilizes static and dynamic signals for next-generation intelligent interface systems.

## 5. Acknowledgments

This work was supported by Dr Min Yu's Imperial College Research Fellowship (ICRF).

## References

[1] Chen S, Zhou Z, Hou K, et al. Artificial organic afferent nerves enable closed-loop tactile feedback for intelligent robot[J]. Nature Communications, 2024, 15(1): 7056.

[2] Wang P, Xie Z, Xin W, et al. Sensing expectation enables simultaneous proprioception and contact detection in an intelligent soft continuum robot[J]. Nature Communications, 2024, 15(1): 9978.

[3] Yang H, Ding S, Wang J, et al. Computational design of ultra-robust strain sensors for soft robot perception and autonomy[J]. Nature Communications, 2024, 15(1): 1636.

[4] Sankar S, Cheng W Y, Zhang J, et al. A natural biomimetic prosthetic hand with neuromorphic tactile sensing for precise and compliant grasping[J]. Science Advances, 2025, 11(10): eadr9300.

[5] Chen Y, Valenzuela C, Liu Y, et al. Biomimetic artificial neuromuscular fiber bundles with built-in adaptive feedback[J]. Matter, 2025, 8(2).

[6] Yang J S, Chung M K, Yoo J Y, et al. Interference-free nanogap pressure sensor array with high spatial resolution for wireless human-machine interfaces applications[J]. Nature Communications, 2025, 16(1): 2024.

[7] Wang J, Lin W, Chen Z, et al. Smart touchless human–machine interaction based on crystalline porous cages[J]. Nature Communications, 2024, 15(1): 1575.

[8] Zhang J, Yao H, Mo J, et al. Finger-inspired rigid-soft hybrid tactile sensor with superior sensitivity at high frequency[J]. Nature communications, 2022, 13(1): 5076.

[9] Bai N, Xue Y, Chen S, et al. A robotic sensory system with high spatiotemporal resolution for texture recognition[J]. Nature Communications, 2023, 14(1): 7121.

[10] Zhuo S, Song C, Rong Q, et al. Shape and stiffness memory ionogels with programmable pressure-resistance response[J]. Nature Communications, 2022, 13(1): 1743.

[11] Qin J, Yin L J, Hao Y N, et al. Flexible and stretchable capacitive sensors with different microstructures[J]. Advanced Materials, 2021,

33(34): 2008267.

[12] Cao M, Su J, Fan S, et al. Wearable piezoresistive pressure sensors based on 3D graphene[J]. Chemical Engineering Journal, 2021, 406: 126777.

[13] Qiu Y, Wang F, Zhang Z, et al. Quantitative softness and texture bimodal haptic sensors for robotic clinical feature identification and intelligent picking[J]. Science Advances, 2024, 10(30): eadp0348.

[14] Huang X, Ma Z, Xia W, et al. A high-sensitivity flexible piezoelectric tactile sensor utilizing an innovative rigid-in-soft structure[J]. Nano Energy, 2024, 129: 110019.

[15] Niu H, Li H, Li N, et al. Fringing-Effect-Based Capacitive Proximity Sensors[J]. Advanced Functional Materials, 2024, 34(51): 2409820.

[16] Cheng T, Shao J, Wang Z L. Triboelectric nanogenerators[J]. Nature Reviews Methods Primers, 2023, 3(1): 39.

[17] Kim W G, Kim D W, Tcho I W, et al. Triboelectric nanogenerator: Structure, mechanism, and applications[J]. ACS nano, 2021, 15(1): 258-287.

[18] Luo J, Gao W, Wang Z L. The triboelectric nanogenerator as an innovative technology toward intelligent sports[J]. Advanced materials, 2021, 33(17): 2004178.

[19] Liu Y, Liu W, Wang Z, et al. Quantifying contact status and the air-breakdown model of charge-excitation triboelectric nanogenerators to maximize charge density[J]. Nature communications, 2020, 11(1): 1599.

[20] Yu Y, Gao Q, Zhang X, et al. Contact-sliding-separation mode triboelectric nanogenerator[J]. Energy & Environmental Science, 2023, 16(9): 3932-3941.

[21] Xia K, Wu D, Fu J, et al. Tunable output performance of triboelectric nanogenerator based on alginate metal complex for sustainable operation of intelligent keyboard sensing system[J]. Nano Energy, 2020, 78: 105263.

[22] Wu H, He W, Shan C, et al. Achieving remarkable charge density via self-polarization of polar high-k material in a charge-excitation triboelectric nanogenerator[J]. Advanced Materials, 2022, 34(13): 2109918.

[23] Xia K, Xu Z. Double-piezoelectric-layer-enhanced triboelectric nanogenerator for bio-mechanical energy harvesting and hot airflow monitoring[J]. Smart Materials and Structures, 2020, 29(9): 095016.

[24] Fu J, Xia K, Xu Z. Double helix triboelectric nanogenerator for self-powered weight sensors[J]. Sensors and Actuators A: Physical, 2021, 323: 112650.

[25] Liu W, Wang Z, Wang G, et al. Integrated charge excitation triboelectric nanogenerator[J]. Nature communications, 2019, 10(1): 1426.

[26] Xia K, Zhu Z, Zhang H, et al. Painting a high-output triboelectric nanogenerator on paper for harvesting energy from human body motion[J]. Nano Energy, 2018, 50: 571-580.

[27] Xia K, Zhu Z, Zhang H, et al. Milk-based triboelectric nanogenerator on paper for harvesting energy from human body motion[J]. Nano Energy, 2019, 56: 400-410.

[28] Xia K, Wu D, Fu J, et al. A high-output triboelectric nanogenerator based on nickel–copper bimetallic hydroxide nanowrinkles for self-powered wearable electronics[J]. Journal of Materials Chemistry A, 2020, 8(48): 25995-26003.

[29] Hasan M A M, Zhu W, Bowen C R, et al. Triboelectric nanogenerators for wind energy harvesting[J]. Nature Reviews Electrical Engineering, 2024, 1(7): 453-465.

[30] Zhu M, Zhu J, Zhu J, et al. Bladeless wind turbine triboelectric nanogenerator for effectively harvesting random gust energy[J]. Advanced Energy Materials, 2024, 14(33): 2401543.

[31] Wang K, Yao Y, Liu Y, et al. Self-powered system for real-time wireless monitoring and early warning of UAV motor vibration based on triboelectric nanogenerator[J]. Nano Energy, 2024, 129: 110012.

[32] Zhang H, Chen Y, Deng Z, et al. A high-output performance disc-shaped liquid-solid triboelectric nanogenerator for harvesting omnidirectional ultra-low-frequency natural vibration energy[J]. Nano Energy, 2024, 121: 109243.

[33] Cao Y, Su E, Sun Y, et al. A Rolling-Bead Triboelectric Nanogenerator for Harvesting Omnidirectional Wind-Induced Energy toward

Shelter Forests Monitoring[J]. Small, 2024, 20(10): 2307119.

[34] Zhang Z, Yu L, Xia Q, et al. An air-triggered contact-separation rotating triboelectric nanogenerator based on rotation-vibration-pressure conversion[J]. Energy Conversion and Management, 2024, 314: 118663.

[35] Xia K, Yu M. Highly robust and efficient metal-free water cup solid–liquid triboelectric nanogenerator for water wave energy harvesting and ethanol detection[J]. Chemical Engineering Journal, 2025, 503: 157938.

[36] Xia K, Xu Z, Hong Y, et al. A free-floating structure triboelectric nanogenerator based on natural wool ball for offshore wind turbine environmental monitoring[J]. Materials Today Sustainability, 2023, 24: 100467.

[37] Xia K, Fu J, Xu Z. Multiple-frequency high-output triboelectric nanogenerator based on a water balloon for all-weather water wave energy harvesting[J]. Advanced Energy Materials, 2020, 10(28): 2000426.

[38] Noor A, Sun M, Zhang X, et al. Recent advances in triboelectric tactile sensors for robot hand[J]. Materials Today Physics, 2024: 101496.

[39] Chen S, Pang Y, Cao Y, et al. Soft robotic manipulation system capable of stiffness variation and dexterous operation for safe human–machine interactions[J]. Advanced Materials Technologies, 2021, 6(5): 2100084.

[40] Zhu M, Sun Z, Zhang Z, et al. Haptic-feedback smart glove as a creative human-machine interface (HMI) for virtual/augmented reality applications[J]. Science Advances, 2020, 6(19): eaaz8693.

[41] Zhu M, Sun Z, Lee C. Soft modular glove with multimodal sensing and augmented haptic feedback enabled by materials' multifunctionalities[J]. ACS nano, 2022, 16(9): 14097-14110.

[42] Chen T, Dai Z, Liu M, et al. 3D multimodal sensing and feedback finger case for immersive dual-way interaction[J]. Advanced Materials Technologies, 2024, 9(5): 2301681.

[43] Bhatta T, Sharma S, Shrestha K, Shin Y, Seonu S, Lee S, Kim D, Sharifuzzaman M, Rana S, Park J. Siloxene/PVDF composite nanofibrous membrane for high-performance triboelectric nanogenerator and self-powered static and dynamic pressure sensing applications. Advanced Functional Materials, 2022, 32(25): 2202145.

[44] Fu X, Dong J, Li L, et al. Fingerprint-inspired dual-mode pressure sensor for robotic static and dynamic perception[J]. Nano Energy, 2022, 103: 107788.

[45] Wei X, Li H, Yue W, et al. A high-accuracy, real-time, intelligent material perception system with a machine-learning-motivated pressure-sensitive electronic skin[J]. Matter, 2022, 5(5): 1481-1501.

[46] Lin L, Xie Y, Wang S, et al. Triboelectric active sensor array for self-powered static and dynamic pressure detection and tactile imaging[J]. ACS nano, 2013, 7(9): 8266-8274.

[47] Li X, Wang J, Liu Y, et al. Lightweight and strong cellulosic triboelectric materials enabled by cell wall nanoengineering[J]. Nano Letters, 2024, 24(10): 3273-3281.

[48] Xia K, Yu M, Luo Y, et al. All-foam intrinsic triboelectric static and dynamic pressure sensor with a standardized DC/AC measurement method for industrial robots[J]. Nano Energy, 2025: 110953.

[49] Wang S, Chen Z, Zhou H, et al. Microporous MXene/Polyurethane Gels Derived from Iron Foam Templates for Ultra-High Stable Pressure Sensors and Triboelectric Nanogenerators[J]. ACS Applied Electronic Materials, 2024, 6(5): 3491-3500.

[50] Zhang H, Li H, Li Y. Biomimetic electronic skin for robots aiming at superior dynamic-static perception and material cognition based on triboelectric-piezoresistive effects[J]. Nano Letters, 2024, 24(13): 4002-4011.

# Supplementary Information

## Intrinsic static/dynamic triboelectric pressure sensor for continuous and event-triggered control

Kequan Xia[1], Song Yang[1], Jianguo Lu[2], Min Yu[1*]

[1]Department of Mechanical Engineering, Imperial College London, SW7 2AZ, London, United Kingdom;

[2]State Key Laboratory of Silicon and Advanced Semiconductor Materials, School of Materials Science and Engineering, Zhejiang University, Hangzhou 310058, China.

*Corresponding author: m.yu14@imperial.ac.uk.

**Supplementary Note 1. Material characterization of the triboelectric layers used in the iSD Sensor.**

**Supplementary Figure S2(a, b)** show the macroscopic and microscopic morphology of the ePTFE film. The material demonstrates good flexibility, as evidenced by its ability to undergo significant bending without structural damage. The lower insets show SEM images of the ePTFE film at 1,000× and 10,000× magnification, revealing its multilayered porous architecture composed of interconnected nano-fibrils. This nanostructured topology contributes to enhanced surface area and charge trapping capability, which are crucial for improving triboelectric performance. **Figure S2(c)** displays the elastic conductive sponge electrode with a size of 4 cm × 4 cm. The sponge provides mechanical compliance and electrical conductivity, supporting conformal contact under compression. **Figure S2(d)** confirms that both sides of the conductive sponge are electrically conductive, as demonstrated by resistance testing (~3.7 Ω). The adhesive substrate attached to the sponge aids in structural integration and ease of assembly in the sensor configuration.

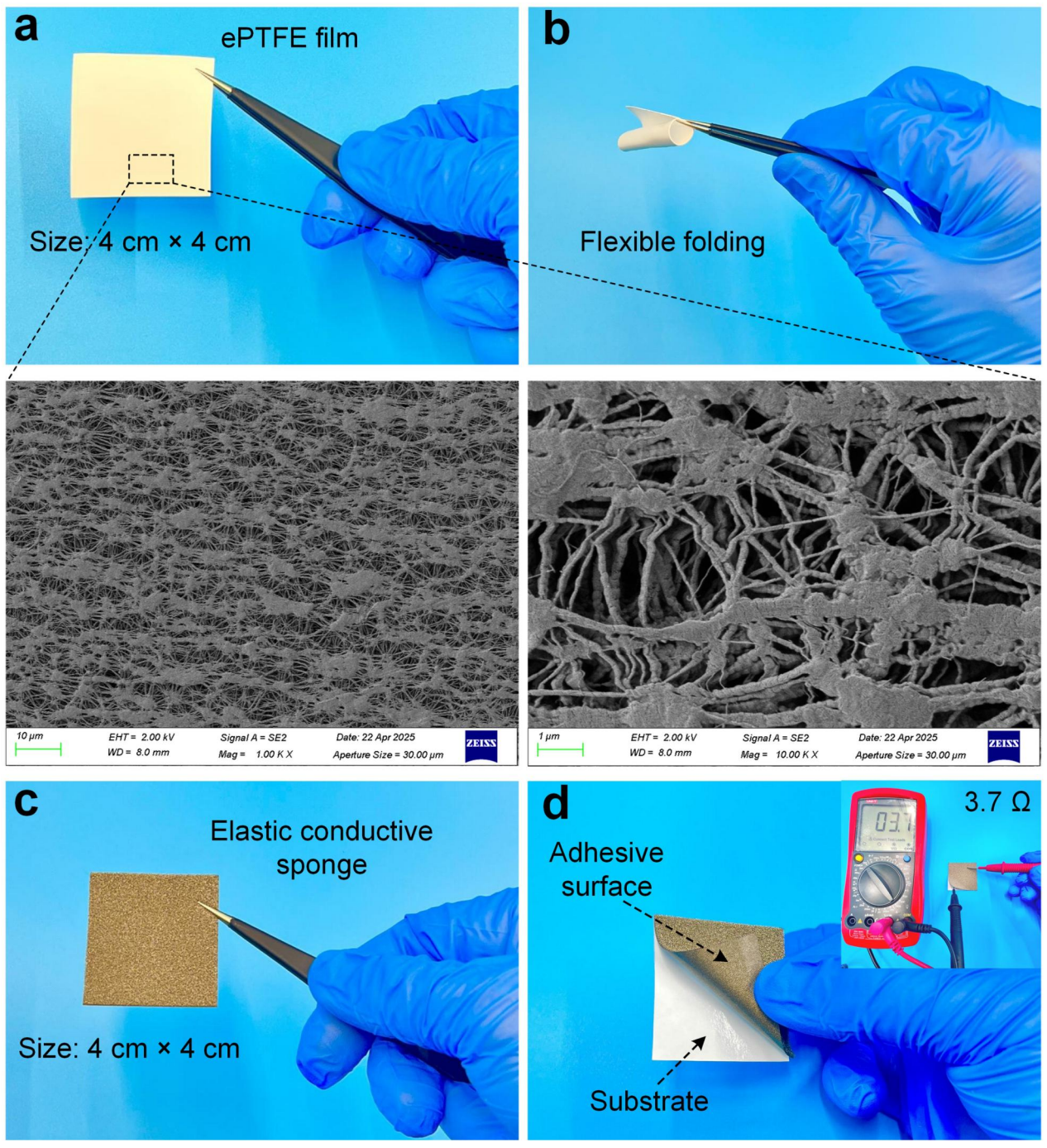

**Figure S1. Physical characterization of triboelectric materials used in the iSD Sensor.** (a) Photograph of the ePTFE film (4 cm × 4 cm). (b) Folding demonstration and corresponding SEM images showing porous nanofibrous structure at low and high magnification. (c) Elastic conductive sponge (4 cm × 4 cm). (d) Confirmation of double-sided conductivity via multimeter resistance measurement (~3.7 Ω) and illustration of adhesive substrate.

**Supplementary Note 2. Experimental setup for characterizing the output performance of the iSD Sensor.**

The experimental system used to characterize the iSD Sensor's static and dynamic pressure responses is shown in **Supplementary Figure S2**. A sinusoidal or square-wave signal was generated using a signal generator, amplified via a broadband power amplifier, and fed to an electrodynamic shaker (exciter) to impose vertical periodic loading. The iSD Sensor was mounted on a rigid platform above a digital pressure gauge, allowing real-time monitoring of applied pressure. The sensor output was simultaneously recorded using a Keithley 6517B electrometer and a digital oscilloscope (Tektronix TBS 2000 series). A thermohygrometer was used to monitor environmental temperature and humidity during all measurements. To ensure precise alignment of the sensor with the pressure axis, a bubble gradienter was positioned alongside the fixture.

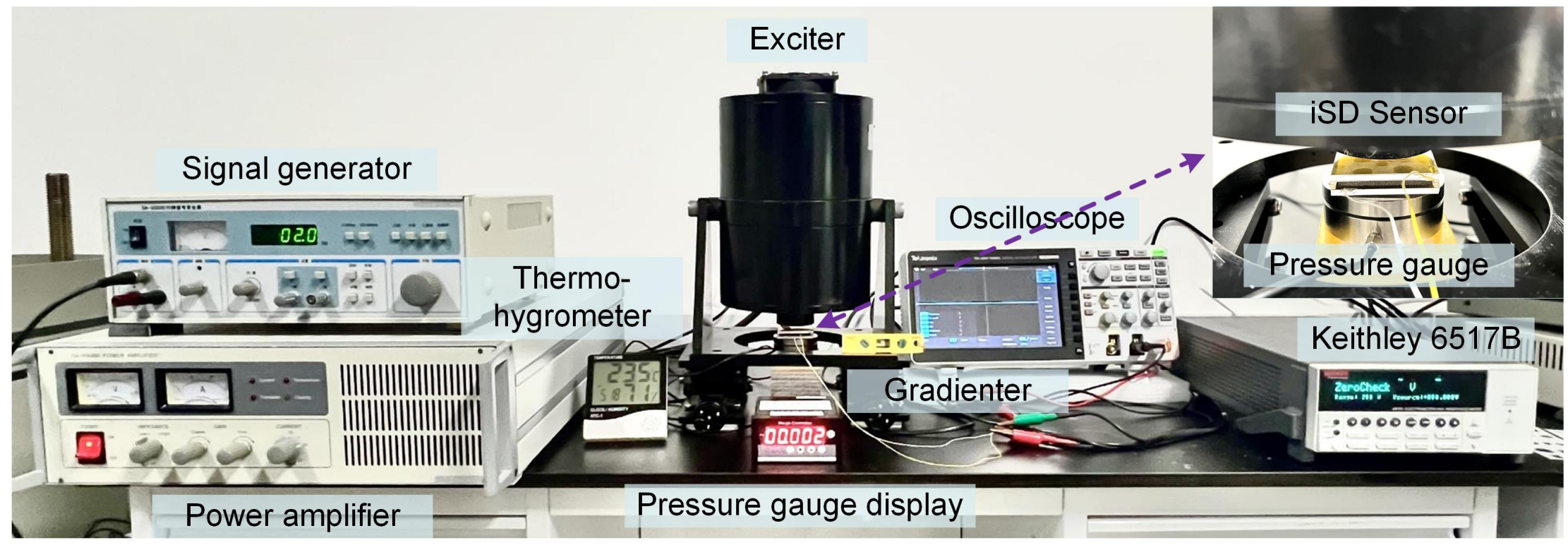

**Figure S2. Experimental setup for testing the iSD Sensor.** A signal generator, power amplifier, and exciter apply controlled pressure, while output signals are measured by a Keithley 6517B electrometer and oscilloscope. A pressure gauge monitors the applied force, and a thermohygrometer tracks environmental conditions.

**Supplementary Note 3. Effect of charge excitation (CE) on signal performance of the iSD Sensor.**

To evaluate the effectiveness and limitations of the charge excitation (CE) strategy, we monitored the real-time voltage output of the iSD Sensor under both positive charge excitation (PCE) and reverse charge excitation (RCE) modes. As shown in **Supplementary Figure S3(a, b)**, both CE modes significantly increased the sensor's baseline signal amplitude, confirming the enhanced surface charge density achieved by the voltage modulation circuit (VMC). However, the voltage waveforms exhibit notable instability and oscillations over time, which may hinder precise pressure sensing in practical applications. These results suggest that while CE effectively boosts triboelectric performance, its use should be restricted to pre-conditioning. For sensing purposes, the VMC should be removed after CE treatment, and the iSD Sensor should operate independently after its pre-excited state to ensure stable and reliable signal output.

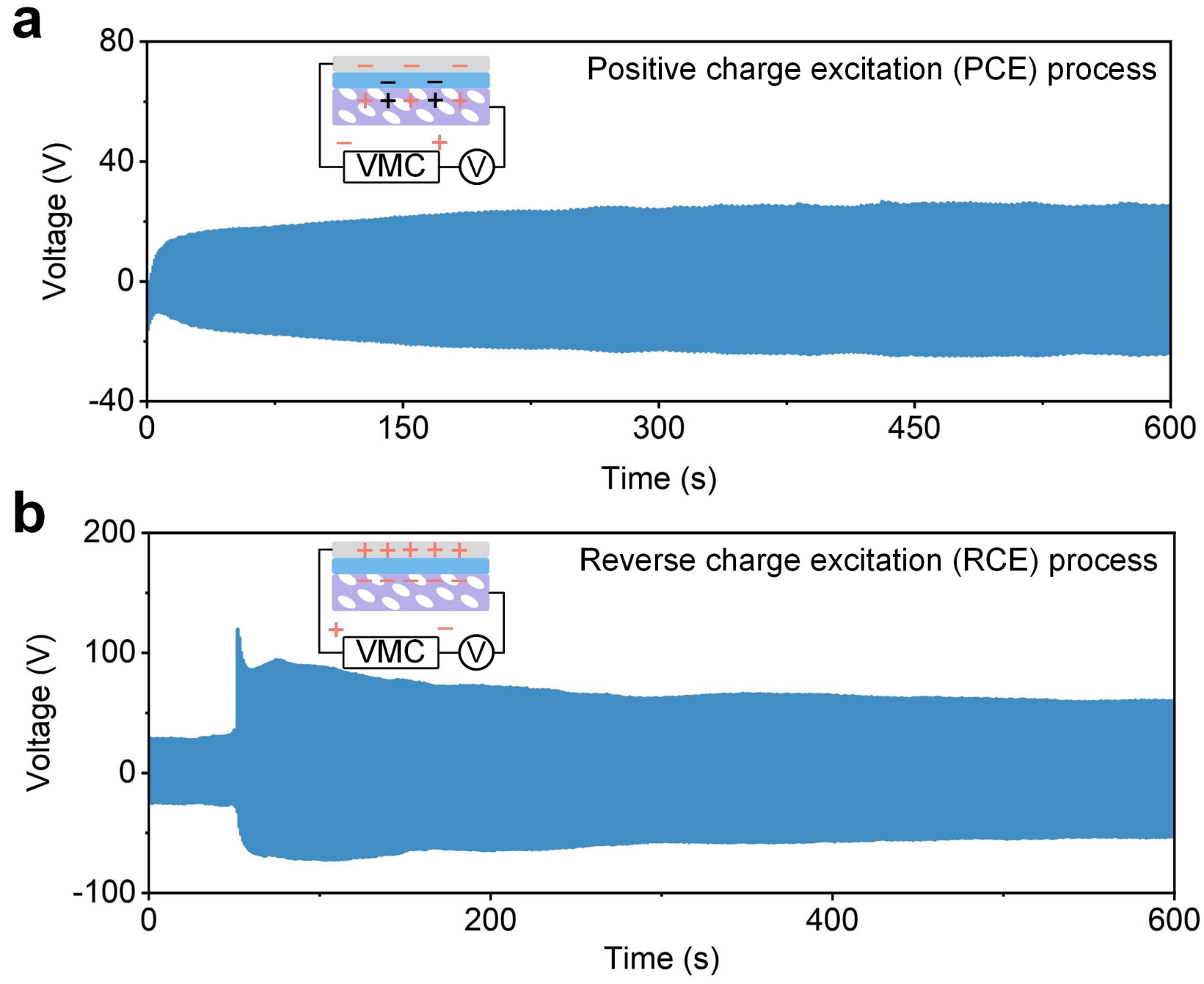

**Figure S3. Charge excitation significantly boosts the output signal of the iSD Sensor but induces waveform instability during operation.** (a) Voltage output of the iSD Sensor under positive charge excitation (PCE), showing enhanced baseline signal with noticeable fluctuations over time. (b) Voltage output under reverse charge excitation (RCE), also exhibiting increased signal amplitude alongside waveform instability.

**Supplementary Note 4. Long-term charge retention and output stability of the iSD Sensor.**

To evaluate the long-term stability of the iSD Sensor, its output performance was monitored after being stored under ambient conditions for up to three weeks without re-excitation. As shown in **Supplementary Figure S4**, the device consistently generated stable, periodic voltage signals at each time point (initial state, after 1 week, 2 weeks, and 3 weeks), indicating excellent output retention over time. This performance stability is primarily attributed to the sensor's intrinsic static pressure sensing mode, in which the triboelectric layers remain in a micro-contact configuration even without external force. Additionally, the encapsulated structure of the sensor effectively isolates the triboelectric interface from environmental influences such as humidity and air flow, thus preventing triboelectric charge dissipation and ensuring robust long-term signal reliability. These results confirm the iSD Sensor's suitability for practical applications requiring sustained, low-maintenance operation.

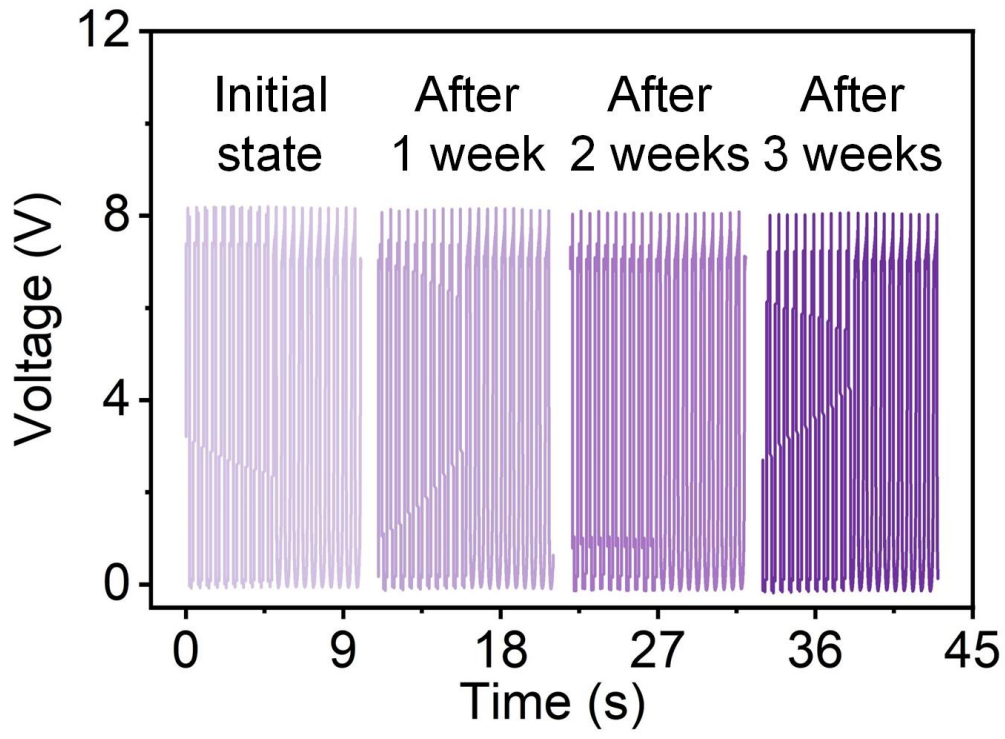

**Figure S4. Long-term stability test of the iSD Sensor.** Output voltage signals recorded under identical dynamic excitation conditions at the initial state, and after 1 week, 2 weeks, and 3 weeks of ambient storage.

**Supplementary Note 5. Theoretical model for static pressure sensing in the iSD Sensor.**

This section provides a theoretical model for understanding the output behavior of the iSD Sensor under static pressure, based on electrostatic principles and mechanical deformation of the triboelectric layers. **Supplementary Figure 5** schematically illustrates the working mechanism of the iSD Sensor in its static pressure sensing mode, highlighting the evolution of contact area and dielectric spacing under compressive loading.

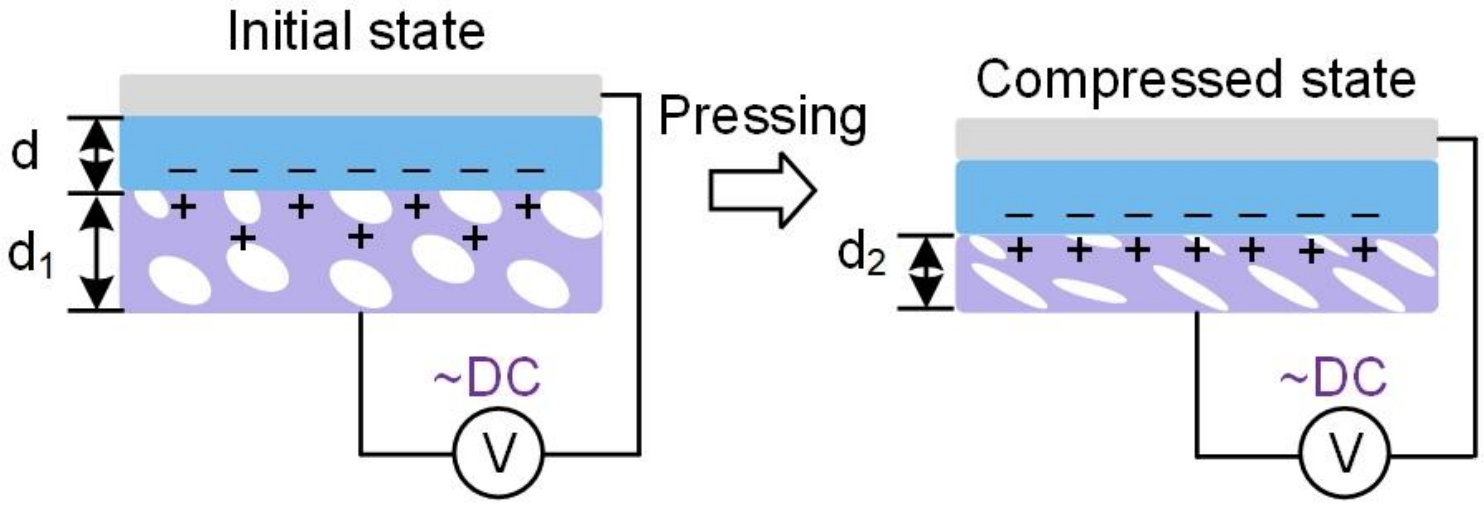

**Figure S5. Schematic illustration of the iSD Sensor under static pressure.**

**(1) Capacitance of the system:** The sensor structure is approximated as a parallel plate capacitor, with the dielectric layer being the ePTFE film:

$$C = \frac{\varepsilon_0 \varepsilon_r A}{d} \tag{1}$$

where $\varepsilon_0$ is the vacuum permittivity, $\varepsilon_r$ is the relative permittivity of the ePTFE layer, $A$ is the effective contact area, and $d$ is the dielectric thickness.

**(2) Output voltage in open-circuit condition:** The $V_{OC}$ is related to the $Q$ and the capacitance:

$$V_{OC} = \frac{Q}{C} = \frac{Qd}{\varepsilon_0 \varepsilon_r A} \tag{2}$$

**(3) Area-pressure relationship:** Due to the compressible nature of the conductive sponge, the effective contact area increases with pressure:

$$A(P) = A_0 + \alpha P \tag{3}$$

where $A_0$ is the initial contact area and $\alpha$ is an empirical area expansion coefficient.

**(4) Thickness-pressure relationship:** The thickness of the ePTFE compresses linearly with pressure:

$$d(P) = d - \beta P \tag{4}$$

where $d_1$ is the initial thickness and $\beta$ is the compressibility constant.

**(5) Voltage-pressure dependency:** Substituting Equations (3) and (4) into (2), we obtain:

$$V_{static}(P) = \frac{Q(d-\beta P)}{\varepsilon_0 \varepsilon_r (A_0 + \alpha P)} \tag{5}$$

This formula captures the nonlinear variation of voltage with pressure.

**(6) Voltage sensitivity to pressure:** The sensitivity $S$ is defined as the derivative of $V_{OC}$ with respect to pressure:

$$S_{static} = \frac{dV_{OC}}{dP} \tag{6}$$

**(7) Expanded derivative for sensitivity analysis:** Differentiating Equation (5) gives:

$$S_{static} = \frac{Q}{\varepsilon_0 \varepsilon_0} \cdot \frac{-\beta(A_0+\alpha P)-\alpha(d-\beta P)}{(A_0+\alpha P)^2} \tag{7}$$

**(8) Limiting case at low pressure:** Assuming $P \to 0$, we obtain the high sensitivity zone:

$$S_{static-max} = \frac{Q}{\varepsilon_0 \varepsilon_0} \cdot \frac{-\beta(A_0)-\alpha d}{{A_0}^2} \tag{8}$$

**(9) Saturation region at high pressure:** When $A(P) \to A_{max}$ and $d(P) \to d_{min}$, sensitivity tends to:

$$S_{static-min} = \frac{Q}{\varepsilon_0 \varepsilon_0} \cdot \frac{-\beta(A_{max})-\alpha d_{min}}{{A_{max}}^2} \tag{9}$$

The model explains the observed multi-zone sensitivity: a steep response in the low-pressure regime due to rapid area expansion, a transition zone with decreasing slope, and a saturation region where both area and thickness change slowly.

**Supplementary Note 6. Optimization of preset separation distance for dynamic pressure sensing.**

To optimize the structural gap for dynamic pressure performance, the iSD Sensor was tested under preset separation distances from 1 mm to 8 mm using sinusoidal loading. As shown in **Supplementary Figure S6(a)**, the preset air gap defines the extent of contact–separation motion. **Figures S6(b–d)** present the output voltage, short-circuit current, and transferred charge at different gaps. While increasing the separation initially boosts signal output due to enhanced deformation and charge transfer, the performance saturates beyond 4 mm, indicating that larger gaps offer limited benefits. Additionally, wider separations demand thicker support structures, which compromise the sensor's responsiveness to small forces. Therefore, a 1 mm gap was chosen as the optimal configuration, balancing sensitivity, resolution, and structural integration for dynamic sensing applications.

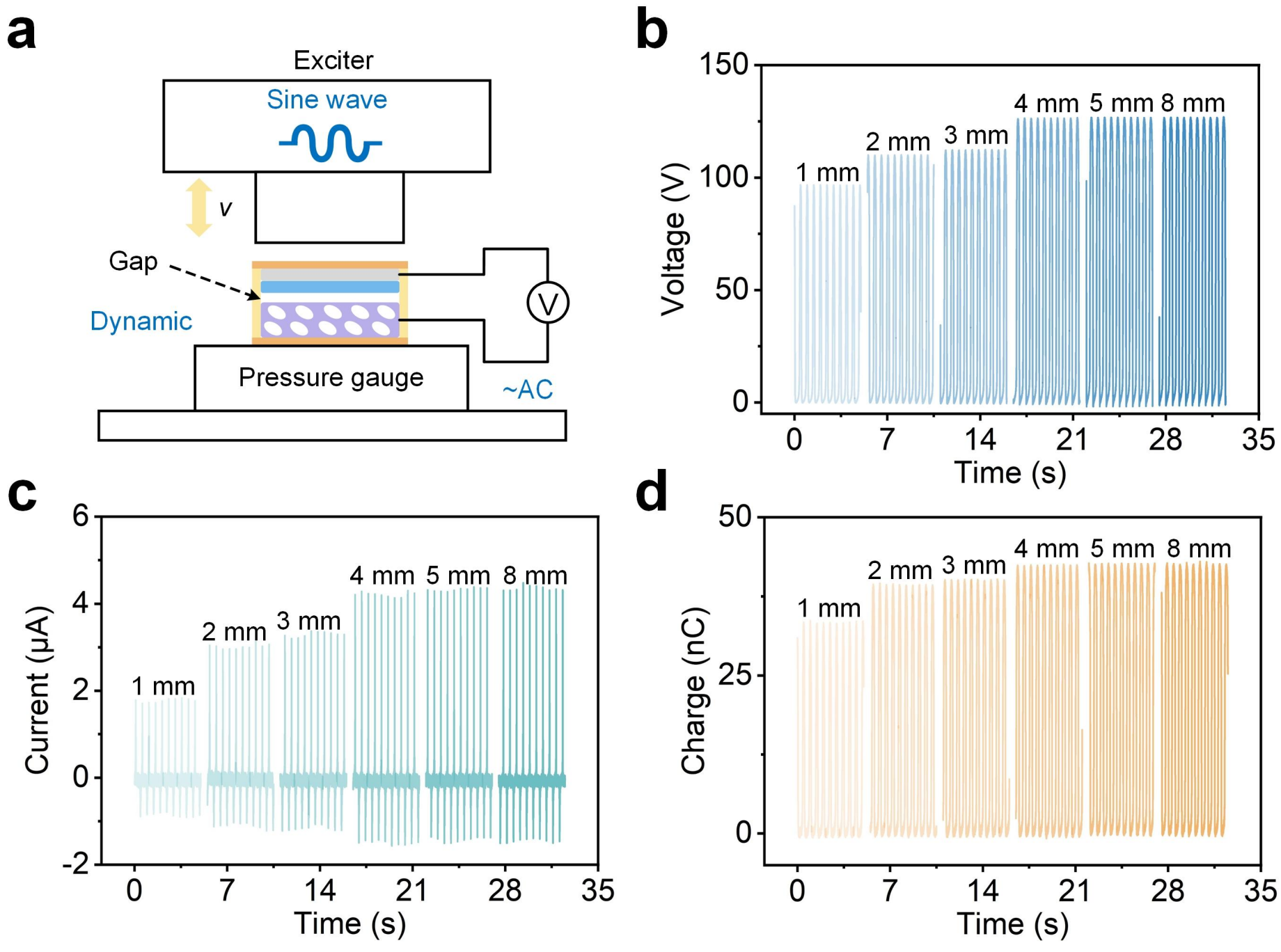

**Figure S6. Dynamic output performance of the iSD Sensor under varied preset separation distances.** (a) Schematic of dynamic sensing structure with adjustable gap. (b) $V_{OC}$, (c) $I_{SC}$, and (d) $Q_{SC}$ under sinusoidal excitation with preset separations of 1–8 mm.

**Supplementary Note 7. Theoretical model for dynamic pressure sensing in the iSD Sensor.**

This section presents a theoretical framework for understanding the dynamic pressure sensing mechanism of the iSD Sensor, as schematically illustrated in **Supplementary Figure S10**. The figure shows the periodic contact–separation process under dynamic mechanical excitation, where the triboelectric layers interact with a fixed maximum separation gap $x$, leading to the generation of AC signals.

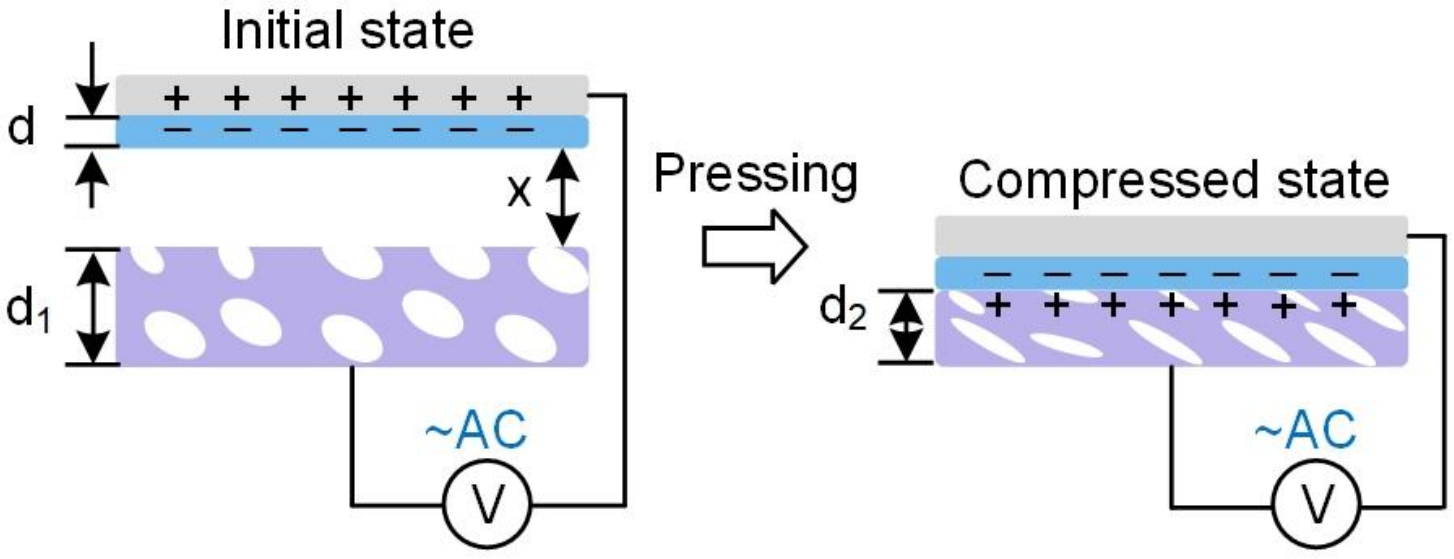

**Figure S7. Schematic illustration of the iSD Sensor under static pressure.**

**(1) Charge generation per cycle:** The total charge transferred in each contact–separation cycle depends on the charge density and contact area:

$$Q_{cycle}(P) = \sigma(P) \cdot A(P) \tag{10}$$

where $\sigma(P)$ is the surface charge density dependent on pressure, and $A(P)$ is the effective contact area.

**(2) Capacitance at maximum separation:** Assuming the system behaves as a parallel-plate capacitor, the minimum capacitance occurs at the maximum separation:

$$C_{min}(P) = \frac{\varepsilon_0 \varepsilon_r A(P)}{x+d} \tag{11}$$

Here, $d$ is the dielectric thickness of the ePTFE layer and $x$ = 1 mm is the fixed separation gap. To better describe the dielectric behavior in dynamic contact-separation systems, we define an effective permittivity $\beta_{eff}$ that considers the series combination of dielectric and air gaps:

$$\frac{1}{\varepsilon_{eff}} = \frac{d}{x+d} \cdot \frac{1}{\varepsilon_r} + \frac{d}{x+d} \cdot \frac{1}{1} \tag{12}$$

This modification enables more accurate modeling of the actual capacitive behavior during fast dynamic transitions.

**(3) Peak voltage per cycle:**

$$V_{peak}(P) = \frac{Q_{cycle}(P)}{C_{min}} = \frac{\sigma(P)(x+d)}{\varepsilon_0 \varepsilon_{eff}} \tag{13}$$

**(4) Empirical voltage-pressure behavior:**

$$V_{dynamic}(P) = V_{max}(1 - e^{-kP}) \tag{14}$$

where $V_{max}$ is the saturation voltage and $k$ is the fitting parameter reflecting pressure sensitivity.

**(5) Pressure-dependent charge density model:**

$$\sigma(P) = \sigma_0(1 - e^{-mP}) \tag{15}$$

where $\sigma_0$ is the maximum charge density and $m$ is a material-dependent constant.

**(6) Dynamic response sensitivity:**

$$S_{dynamic} = \frac{dV_{dynamic}}{dP} = V_{max} \cdot k \cdot e^{-kp} \tag{16}$$

**(7) Initial low-pressure region:**

$$V_{dynamic}(P) \approx V_{max} \cdot kP \tag{17}$$

**(8) Transition region:**

$$S_{dynamic} \downarrow as\ P\ \uparrow \tag{18}$$

**(9) Saturation region:**

$$\lim_{P\to\infty} V(P) = V_{max} \tag{19}$$

This dynamic model distinguishes itself from the static model by incorporating the fixed air gap and the nature of periodic motion. The voltage is primarily controlled by total charge per cycle and maximum separation. By introducing the effective permittivity concept, this model also accounts for layered dielectric effects, offering a reliable foundation for interpreting AC-mode triboelectric sensing performance in fast-response tactile systems.

**Supplementary Note 8. Frequency-dependent performance of the iSD Sensor in dynamic sensing mode.**

To assess the frequency-dependent output characteristics of the iSD Sensor under dynamic pressure sensing conditions, the short-circuit current ($I_{SC}$) and transferred charge ($Q_{SC}$) were measured at excitation frequencies ranging from 2 to 6 Hz under constant pressure amplitude. As shown in **Supplementary Figure S8(a)**, $I_{SC}$ increases steadily with frequency, reaching up to ~10.9 μA at 6 Hz, indicating enhanced charge flow due to more frequent mechanical stimulation. Meanwhile, the corresponding $Q_{SC}$ signals (**Figure S8(b)**) remain nearly constant at ~43.1 nC, demonstrating excellent charge retention and frequency-independent total charge generation. These results confirm the iSD Sensor's stable and robust dynamic sensing performance across a wide frequency range, which is critical for applications involving motion detection, robotic interaction, and bio-mechanical monitoring.

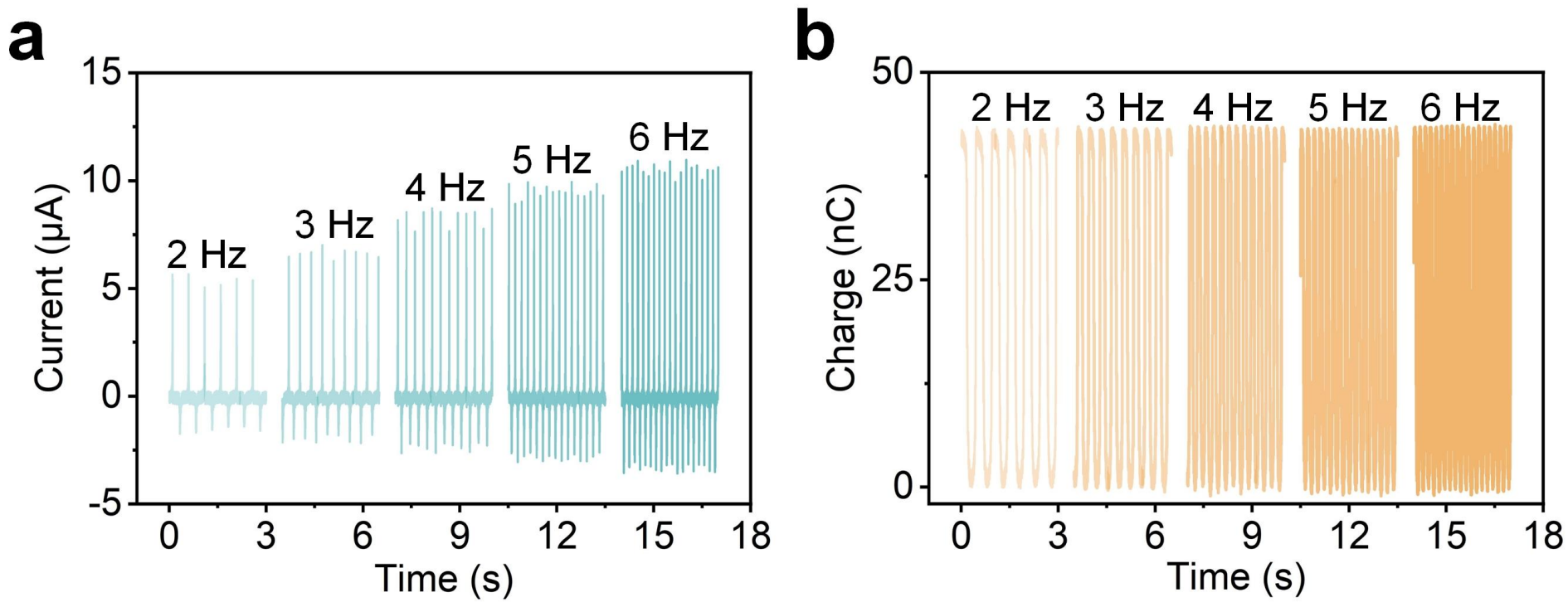

**Supplementary Figure S8. Frequency-dependent electrical performance of the iSD Sensor under dynamic loading.** (a) Short-circuit current ($I_{SC}$) output at frequencies from 2 to 6 Hz, showing increasing peak current with frequency. (b) Transferred charge ($Q_{SC}$) under the same frequency conditions, exhibiting stable charge accumulation across all tested frequencies.

**Supplementary Note 9. Environmental stability of the iSD Sensor under varying environmental conditions.**

**Supplementary Figure S9(a)** shows the contact angles of water droplets on three different positions of the ePTFE film surface, indicating consistent hydrophobicity across the film. The measured contact angles remain stable at approximately 113°, confirming uniform surface wettability essential for maintaining reliable triboelectric performance. To evaluate the environmental robustness of the iSD Sensor, we tested its static and dynamic voltage outputs under different humidity (15–80%) and temperature (20–53 °C) conditions. As shown in **Supplementary Figure S9(b)**, both static and dynamic sensing modes maintain stable voltage outputs across a wide range of relative humidity, indicating the hydrophobicity and encapsulation stability of the triboelectric interface. Furthermore, **Supplementary Figure S9(c)** demonstrates that both sensing modes are resistant to temperature fluctuations, with negligible performance degradation up to 53 °C. These results confirm the sensor's high environmental stability, ensuring reliable operation under ambient variations and supporting its applicability in real-world robotic and wearable scenarios.

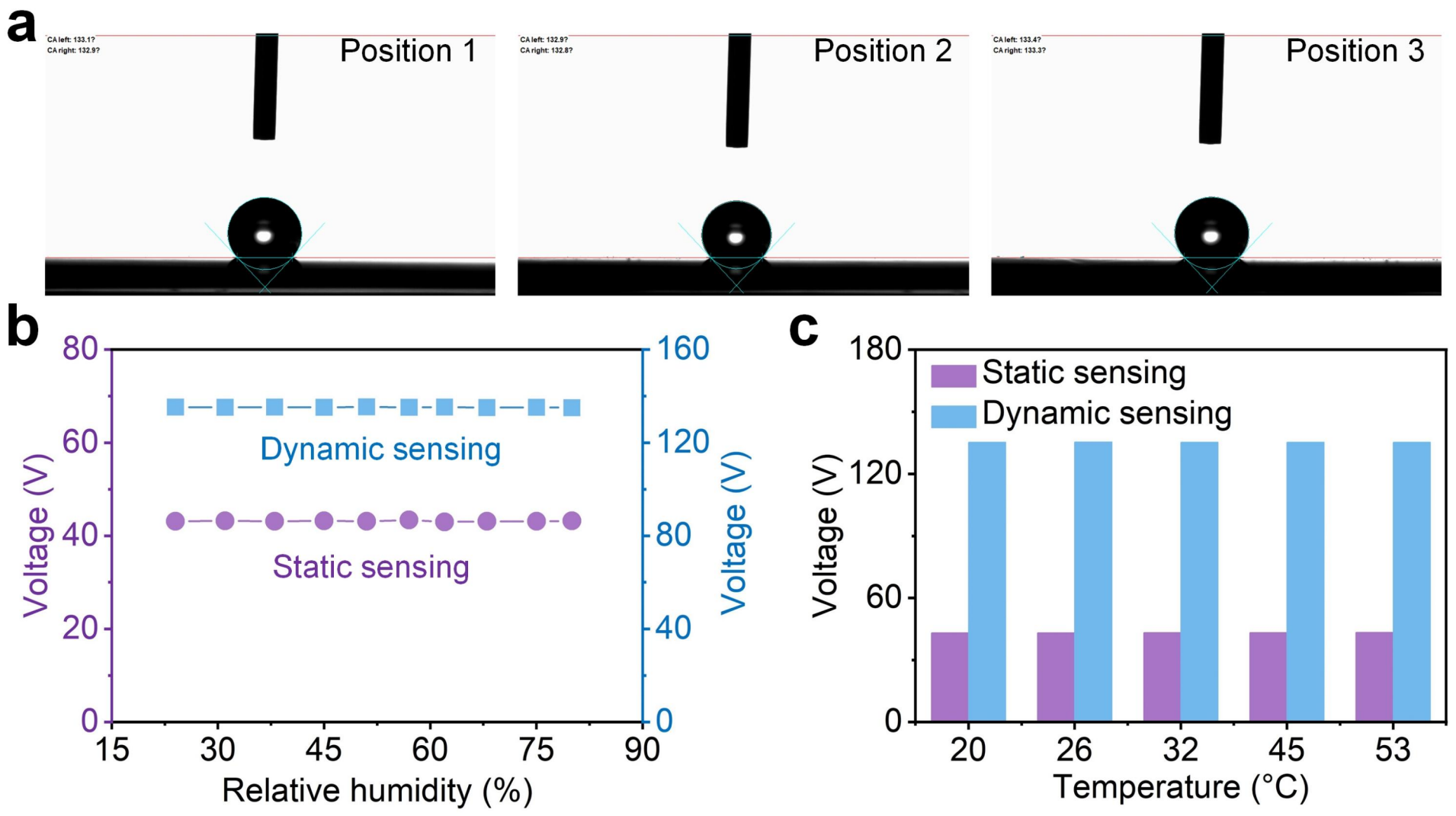

**Figure S9. Environmental robustness of the iSD Sensor.** (a) Contact angles at different positions on the surface of ePTFE film. (b) Static and dynamic voltage outputs under different humidity levels (15% to 85% RH). (c) Static and dynamic outputs under different temperatures (20–53 °C), showing excellent thermal stability.

**Supplementary Note 10. Energy storage behavior of the iSD Sensor under dynamic excitation.**

To evaluate the energy harvesting potential of the iSD Sensor in dynamic conditions, a rectified charging circuit was constructed using a bridge rectifier and parallel storage capacitors, as illustrated in **Supplementary Figure S10(a)**. The iSD Sensor was driven by a programmable exciter applying sinusoidal mechanical loading, and the output was rectified and used to charge capacitors of varying capacitance values. **As shown in Figure S10(b)**, smaller capacitors (e.g., 2.2 μF) reached higher voltages more rapidly, while larger capacitors charged more slowly due to their higher storage capacity. Additionally, **Figure S10(c)** demonstrates the effect of excitation frequency on charging performance. With increasing frequency from 2 to 6 Hz, the charging rate significantly improved, confirming that dynamic response frequency is positively correlated with harvested energy. These results validate the iSD Sensor's capability for real-time energy conversion and support its potential for integration in self-powered systems with auxiliary storage modules.

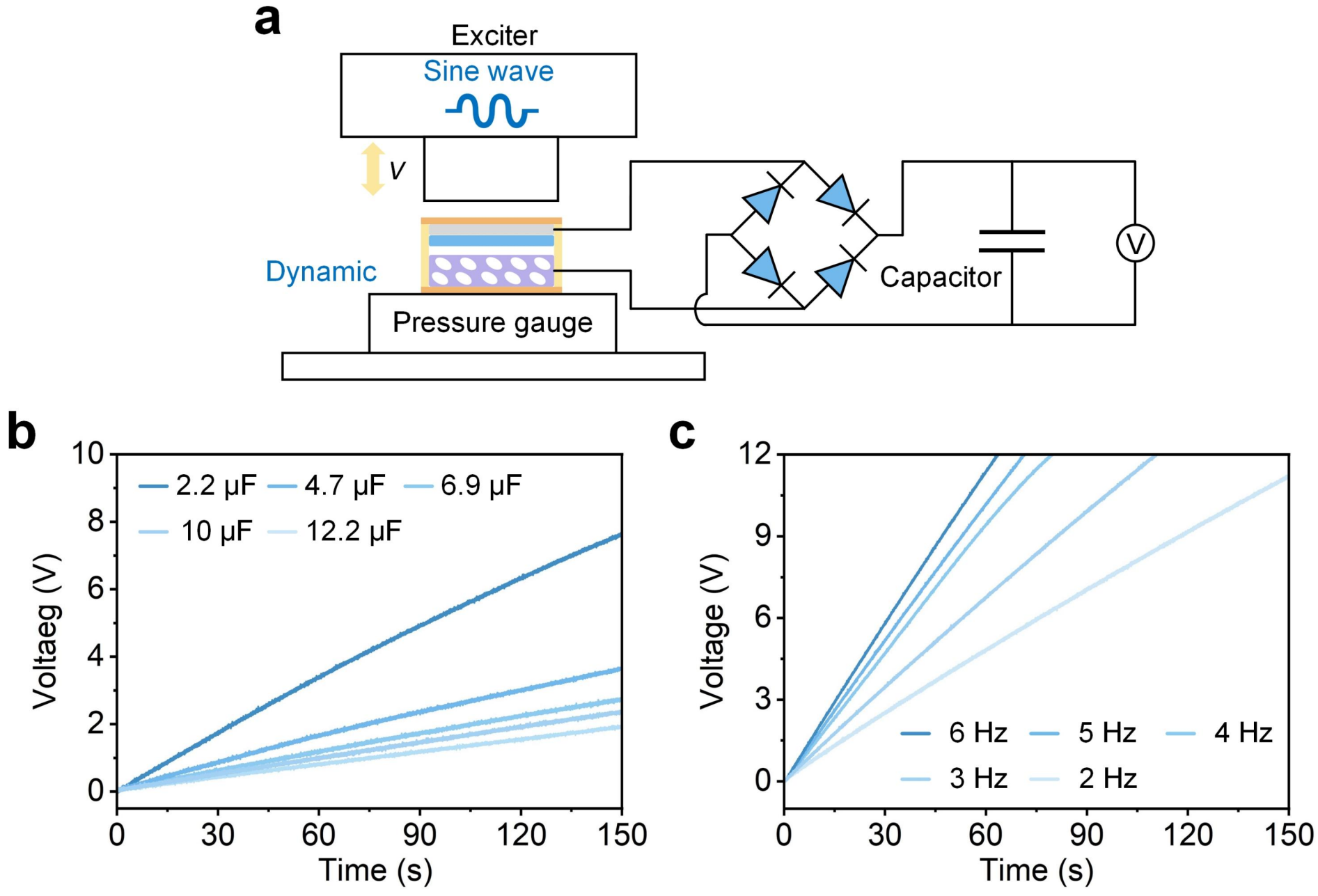

**Figure S10. Energy storage performance of the iSD Sensor under dynamic loading.** (a) Schematic of the energy harvesting setup using a rectifier and capacitor charging circuit. (b) Voltage output curves for capacitors with different capacitances under identical excitation conditions. (c) Charging performance of a fixed capacitor (e.g., 2.7 μF) at varying excitation frequencies (2–6 Hz), showing improved charging rates with frequency increase.

**Supplementary Note 11. Fabrication process of the 3D gradient conductive sponge structure.**

The 3D gradient structure used in the iSD Sensor was fabricated by sequentially stacking multiple layers of conductive sponge with increasing lateral dimensions. As shown in **Supplementary Figure S11**, the base layer is bonded to the substrate, and additional sponge layers are adhered one by one using a thin adhesive layer to ensure mechanical stability and electrical continuity. The stepwise increase in contact area from top to bottom introduces a built-in gradient in compressibility, which is critical to achieving progressive contact under pressure. This design enhances low-pressure sensitivity by enabling early contact at the topmost layer and gradually engaging deeper layers with increasing pressure. The stacking method also provides a robust and reproducible fabrication approach without requiring advanced equipment, making it suitable for scalable production of gradient-structured tactile sensors.

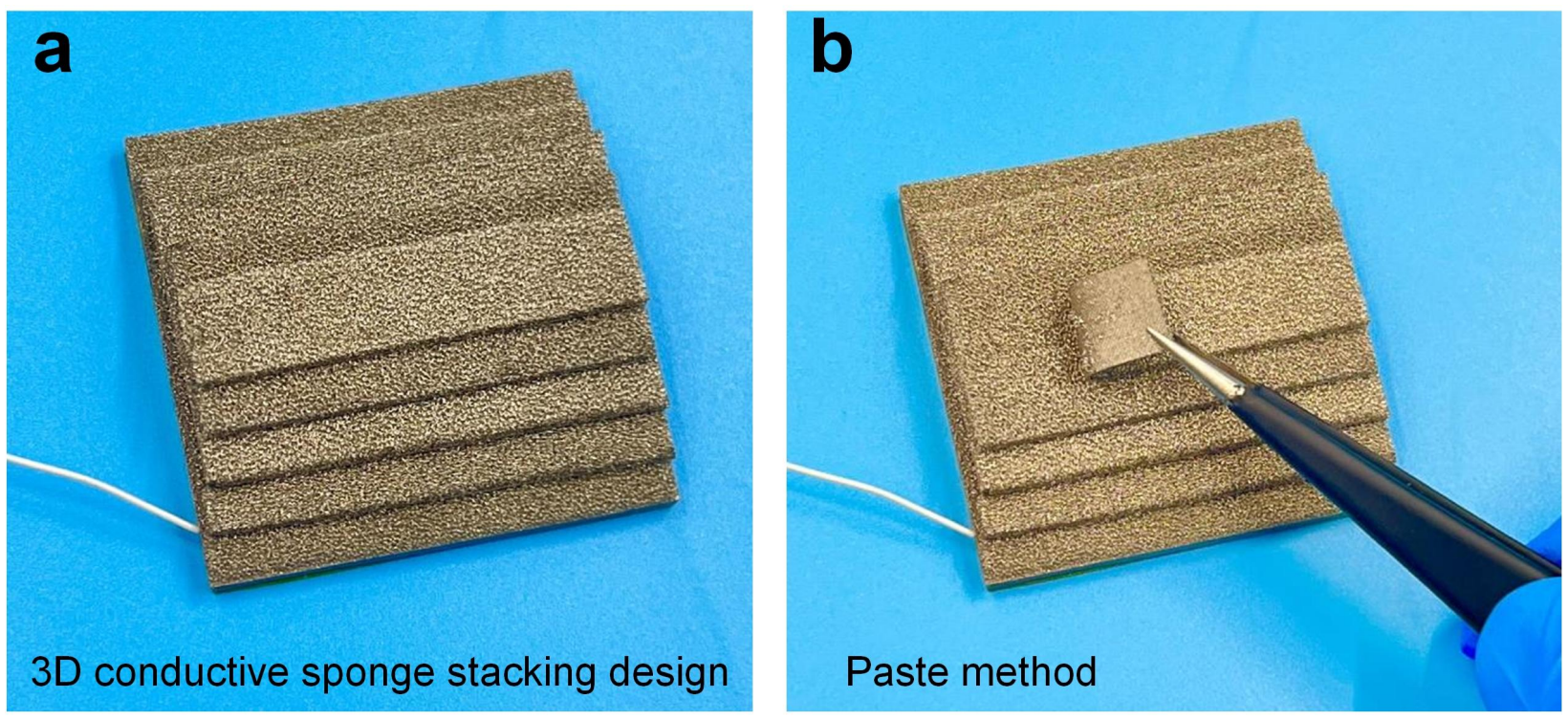

**Figure S11. Fabrication of the 3D gradient conductive sponge structure.** (a) Gradually layered conductive sponge structure forming the 3D gradient architecture. (b) Paste method used to assemble each sponge layer in the stacked configuration.

**Supplementary Note 12. Manual demonstration of static and dynamic pressure modes using the iSD Sensor.**

To further validate the dual-mode signal generation capability of the iSD Sensor under realistic manual operation, we conducted hand-driven experiments simulating typical tactile inputs. **Supplementary Figure S12(a)** shows the experimental setup, where the iSD Sensor is connected to a Keithley 6517B electrometer and a Tektronix oscilloscope to simultaneously record voltage signals under finger-applied pressure. A digital force gauge is used to monitor the applied load in real time.

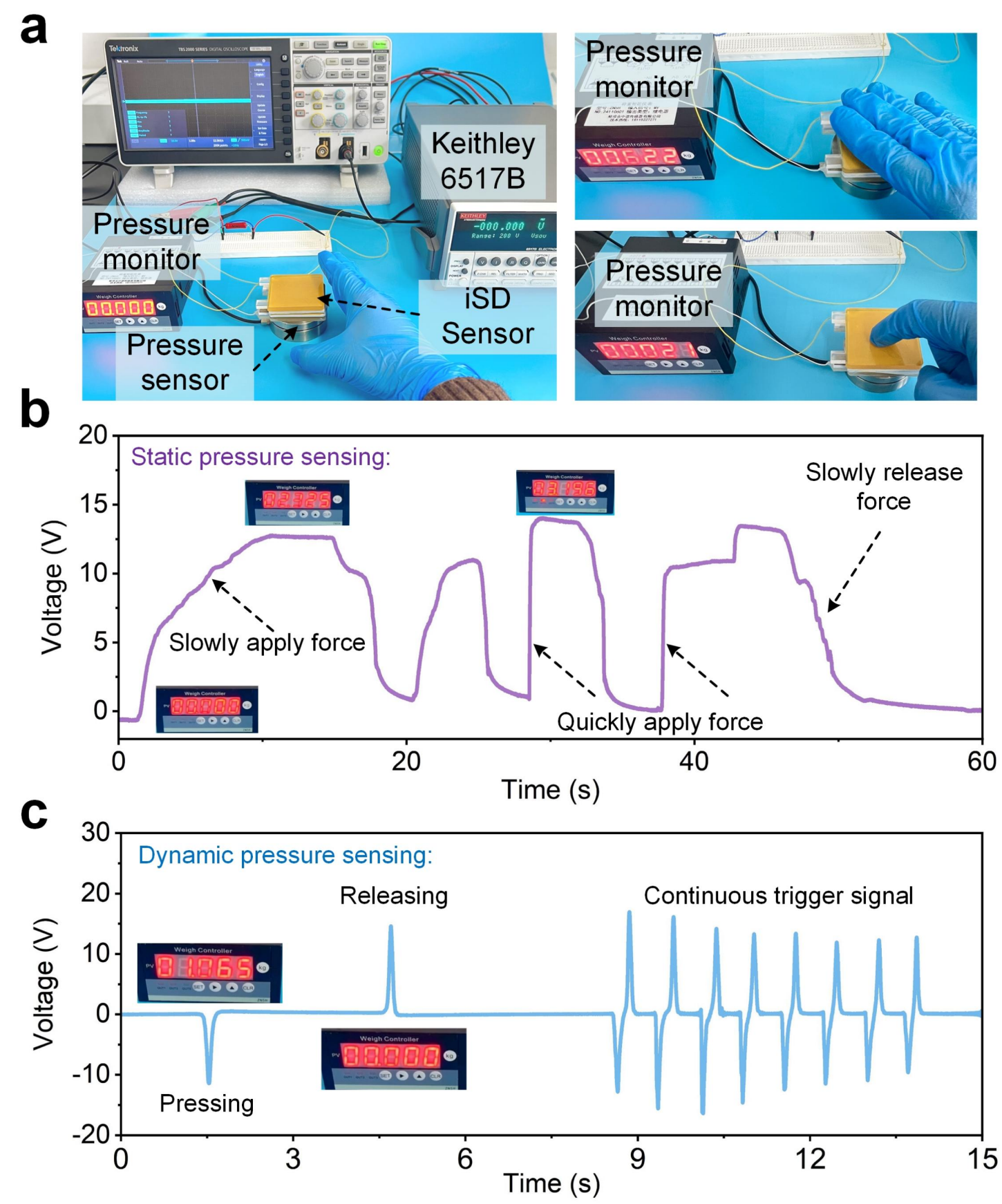

**Figure S12. Manual demonstration of dual-mode pressure response.** (a) Experimental setup showing finger-driven pressure applied to the iSD Sensor with real-time signal recording. (b) Static pressure sensing: gradual or sustained force yields stable, continuous voltage signals. (c) Dynamic pressure sensing: rapid tapping generates distinct bipolar spikes and continuous triggering outputs under repeated excitation.

**As shown in Supplementary Figure S12(b)**, under static pressure input, where the finger presses and holds the sensor, the output voltage rises gradually and maintains a stable plateau until the force is released. Both slow and fast loading scenarios produce sustained signals with amplitudes corresponding to the applied force level. This behavior highlights the iSD Sensor's ability to generate continuous and resolvable DC-like outputs for steady-state pressure, which is critical for proportional control in robotic applications. **In contrast, Supplementary Figure S12(c)** illustrates the sensor's response under dynamic pressure input, where the finger rapidly taps or intermittently presses the sensor. These transient stimuli generate distinct bipolar voltage spikes, with peak amplitudes reflecting the speed and intensity of the motion. Multiple rapid taps produce a train of alternating signals, ideal for fast triggering and sequential command execution. Together, these results demonstrate that the iSD Sensor can effectively distinguish between static and dynamic mechanical inputs based on their signal characteristics. This dual-mode behavior forms the foundation for the subsequent robotic control scheme, where static signals modulate the degree of actuation, while dynamic signals act as triggers for rapid response tasks. The ability to extract both types of information from a single self-powered sensor enables simplified circuitry, compact integration, and responsive control in real-time robotic applications.